\documentclass[11pt]{article}

\usepackage{acl}

\usepackage{times}
\usepackage{latexsym}

\usepackage[T1]{fontenc}

\usepackage[utf8]{inputenc}

\usepackage{microtype}

\usepackage{inconsolata}

\usepackage{graphicx}

\usepackage{amsmath}
\usepackage{amssymb}
\usepackage{mathtools}
\usepackage{amsthm}
\usepackage{url}
\usepackage{enumerate}
\usepackage{fontawesome}
\usepackage{enumitem}
\usepackage{algorithm}
\usepackage{algpseudocode}
\usepackage{multirow}
\usepackage{listings}
\usepackage{arydshln}
\usepackage{adjustbox}
\usepackage{xcolor}
\usepackage{colortbl}
\usepackage[table]{xcolor}
\usepackage{subcaption}
\usepackage{wrapfig}
\usepackage[most]{tcolorbox}
\usepackage{xspace}
\usepackage{placeins}
\usepackage{booktabs} 
\usepackage{multicol}
\usepackage{pifont}

\definecolor{Gainsboro}{rgb}{0.86,0.86,0.86}
\definecolor{nred}{RGB}{196, 38, 11}
\definecolor{ngreen}{RGB}{18, 141, 21}
\definecolor{nblue}{RGB}{93, 143, 240}
\definecolor{norange}{RGB}{242, 212, 102}
\definecolor{nmark}{RGB}{171,223,224}
\definecolor{nyellow}{RGB}{251, 242, 211}
\definecolor{textblue}{RGB}{41, 31, 222}
\newtcolorbox{alprompt}[1]{ boxrule = 1pt, fontupper = \small\tt, fonttitle = \bf\color{black}, arc = 2pt, rounded corners, colframe = black, colbacktitle = white!97!yellow, colback = white!97!yellow, title = #1, }
\newtcolorbox{promptbox}[3][Prompt]{ breakable, colback=black!5!white, arc=5pt, boxrule=0.5pt, fonttitle=\bfseries, title=#1, before upper={\small}, fontupper=\fontfamily{ptm}\selectfont, colframe=#2, label=#3, }

\newtcbox{\hlmarktab}{on line, rounded corners, box align=base, colback=nmark!60,colframe=white,size=fbox,arc=3pt, before upper=\strut, top=-2pt, bottom=-4pt, left=-2pt, right=-2pt, boxrule=0pt}
\newcommand{\marktab}{}
\newcommand{\marktext}[1]{{\hlmarktab{\marktab{#1}}}}

  \newcommand{\headernodot}[1]{\smallskip
  \noindent
  \textbf{#1}}
  \newcommand{\header}[1]{\headernodot{#1.}}

\newcommand{\CommentBlue}[1]{\hfill \textcolor{textblue}{$\triangleright$~\texttt{#1}}}

\newcommand{\ours}{\textsc{CogRE}\xspace}
\newcommand{\reward}{\textsc{Hit@Dict}\xspace}

\newtcbox{\hlbluetab}{on line, rounded corners, box align=base, colback=nblue!60,colframe=white,size=fbox,arc=3pt, before upper=\strut, top=-2pt, bottom=-4pt, left=-2pt, right=-2pt, boxrule=0pt}
\newcommand{\bluetab}{}
\newcommand{\bluetext}[1]{{\hlbluetab{\bluetab{#1}}}}

\newtcbox{\hlshallowbluetab}{on line, rounded corners, box align=base, colback=nblue!10,colframe=white,size=fbox,arc=3pt, before upper=\strut, top=-2pt, bottom=-4pt, left=-2pt, right=-2pt, boxrule=0pt}
\newcommand{\shallowbluetab}{}
\newcommand{\shallowbluetext}[1]{{\hlshallowbluetab{\shallowbluetab{#1}}}}

\newtcbox{\hlyellowtab}{on line, box align=base, colback=norange!50,colframe=white,size=fbox,arc=3pt, before upper=\strut, top=-2pt, bottom=-4pt, left=-2pt, right=-2pt, boxrule=0pt}
\newcommand{\yellow}{}
\newcommand{\yellowtext}[1]{{\hlyellowtab{\yellow{#1}}}}

\newcommand{\xmark}{{\ding{55}}}

\usepackage[capitalize, noabbrev]{cleveref}

\theoremstyle{plain}

\theoremstyle{definition}

\theoremstyle{remark}

\usepackage[textsize=tiny]{todonotes}

\title{\textit{The Answer Lies Within}: \\ Self-Derived Rewards Enable Explainable Relation Extraction}

\author{
    Xinyu Guo\textsuperscript{\normalfont 1},
    Zhengliang Shi\textsuperscript{\normalfont 2},
    Minglai Yang\textsuperscript{\normalfont 1},
    Mahdi Rahimi\textsuperscript{\normalfont 1},
    Mihai Surdeanu\textsuperscript{\normalfont 1}
    \\ 
    \textsuperscript{1}University of Arizona, Tucson, AZ, United States\\
    \textsuperscript{2}Shandong University, Qingdao, China.\\
    \textit{xinyuguo@arizona.edu, msurdeanu@arizona.edu}
}

\begin{document}
\maketitle
\begin{abstract}

Despite their strong general reasoning, large language models still struggle with one-shot relation extraction when no candidate relation labels are given. We identify two failure modes: models often attend to irrelevant tokens rather than to the phrases that actually carry the relation, and they predict relations at a different level of abstraction from the one human annotators expect.
We address both with two components: (1) \textit{\ours, a cognitively inspired reasoning framework} that structures RE into a series of processes mimicking 
human text-processing; 
and (2) \textit{\reward, a reinforcement learning intermediate reward strategy} that 
rewards relation-bearing phrases in the reasoning trace, aligning the reasoning with the target label. The reward is computed against a credit dictionary, a set of high-value phrases extracted automatically from the model's own correct predictions.
%
%
Our experiments show that our framework improves both accuracy and explanation quality by addressing these two challenges. For example, \ours with Qwen2.5-14B-Instruct on One-shot NYT29 achieves 24.65\% F1, surpassing prior reasoning-based designs. Optimizing this approach with RL using \reward further improves performance by +23.46\% points. 
Finally, human evaluation shows that our best model generates relational phrases
closely aligned with gold labels, 
increasing human explanation quality ratings by 54\% (relative). 
\href{https://github.com/xinnyuguo/CogRE-Hit-Dict}{\faGithub \textbf{ }Code}
\end{abstract}


    %
    %
    %

\section{Introduction}\label{sec:introduction}

Relation extraction (RE) aims to identify relations between entity mentions in text~\citep{zelenko2003kernel, bunescu2005shortest} and serves as a fundamental task in high-stakes domains~\citep{adadi2018peeking,goodman2017european}.
However, in such domains explainability is critical;  traditional RE methods based on feature-based models, neural network, or small language models~\citep{kambhatla2004combining,zeng2014relation,soares2019matching, vacareanu2024bestworldspliablegeneralizable} suffer from limited explainability.
While recent works prompt large language models (LLMs) to perform structured reasoning before extracting relations, they typically rely on handcrafted relation labels and descriptions that are expensive to annotate~\citep{li2023revisitinglargelanguagemodels}. 
%
To this end, we study a variant of the realistic one-shot RE setting~\citep{alam2024realisticfewshotrelationextraction} that demands both generalization and explainability. \textit{Formally, given a support sentence for each relation and no explicit relation labels, models are required to both extract relations from test sentences and generate explanations for their predictions.}

\begin{table}[t!]
\captionsetup{skip=4pt}
\centering
\renewcommand\arraystretch{1}
\begin{adjustbox}{width=1\columnwidth,center}

\begin{tabular}{lp{0.89\linewidth}}
\toprule
\textbf{Label} & 
\cellcolor{gray!10}{\yellowtext{of support sentence:}} \quad \textit{per:alternate\_name}\\ 
&   \bluetext{of test sentence:} \quad \cellcolor{gray!20}{\textit{no\_relation}}\\
\midrule
\textbf{Support Sentence} & \textbf{Gadahn} is also known as \textbf{Azzam al-Amriki}. \\
\textbf{Test Sentence} & \textbf{Arcandor} was known as Karstadt in \textbf{2000}. \\
\midrule
\textbf{LLMs Prediction} & These two sentences convey the same relation: an entity is also known by another name. 
\\
&
\cellcolor{gray!20}{Test sentence: \bluetext{\textit{per:alternate\_name}} \textcolor{red}{\xmark}.}
\\
\bottomrule
\end{tabular}
\end{adjustbox}
\caption{An example of attention misalignment for the task of predicting whether the test sentence has the same relation label as a provided support sentence. Bold text marks the target entity pair. The model predicts the same relation based on superficial lexical overlap (e.g., ``known as'') despite an incompatible target entity pair.}\label{tab:error_pattern_1}
\vspace{-4pt}
\end{table}



However, our pilot experiments show that, despite the strong language understanding and reasoning capabilities of LLMs~\citep{gao2023confuciusiterativetoollearning, luo2024improvemathematicalreasoninglanguage,ahn2024largelanguagemodelsmathematical}, they still perform poorly in realistic one-shot RE with two common failure patterns:
(1) \textit{Attention Misalignment}: models fail to focus on the semantics that truly convey the relation. As in the example shown in Table~\ref{tab:error_pattern_1}, models are often distracted by irrelevant tokens in test sentences when attempting to mimic the relation expressed in the support sentences.
(2) \textit{Granularity Mismatch}: models fail to align with the granularity of human annotation schemas. Without predefined relation labels, they struggle to distinguish fine-grained relations (e.g., city-level \textit{vs.}\ country-level headquarters relations).

To address these problems, we propose a structured reasoning framework inspired by human cognition (\ours) and use reinforcement learning (RL) with a \reward reward to bring LLM reasoning in line with human annotation schemas.

\begin{figure*}[t!]
    \centering
    \includegraphics[width=1\linewidth]{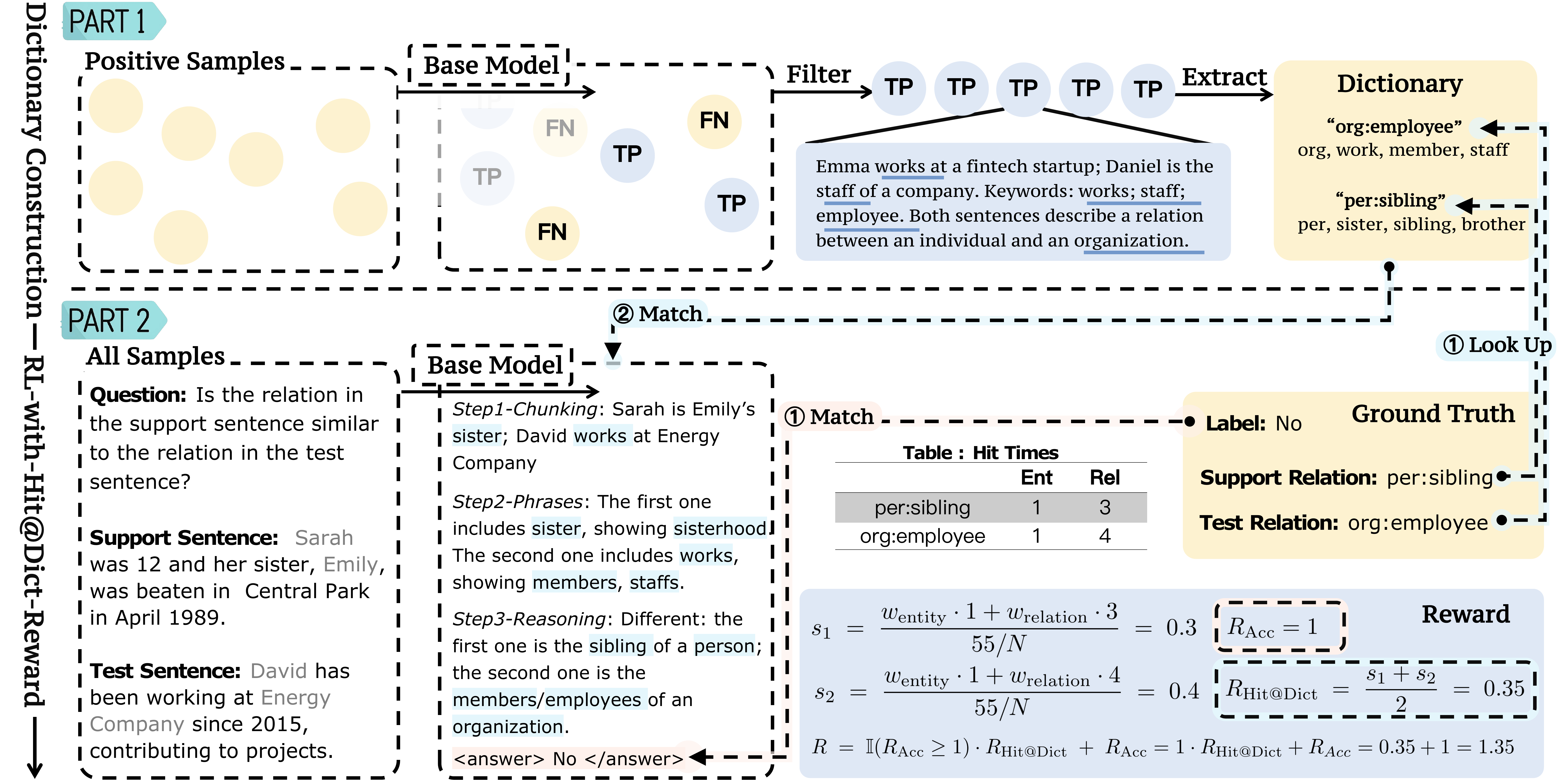}
    \caption{Framework overview. \marktext{\textit{Part 1 - Dictionary Construction}}: We apply a base model to positive samples, retain true positive (TP) instances, then use GPT-4o or Qwen2.5-14B-Instruct to extract relational phrases from their explanations to construct a relational phrase dictionary.
    \marktext{\textit{Part 2 - Reinforcement Learning with \reward}}: During RL training, the model’s stepwise CogRE outputs on all training samples are scored by accuracy (answers) and \reward (explanations). To compute \reward, we first look up relational phrases from the dictionary for both the support and test relations. These phrases are then matched against the LLM outputs to obtain hit statistics (Hit Times Table), and \reward is computed as a normalized hit rate (Section~\ref{sec:hit_reward}).}
    \label{fig:main-figure}
    \vspace{-10pt} 
\end{figure*}

Specifically, CogRE mimics how people process text: comprehension does not come from storing input word by word~\citep{miller1956magical}, but from a construction–integration process that yields a coherent logical chain~\citep{kintsch1988role}.
Inspired by this, we decompose RE into three steps (shown in Part 2 of Figure~\ref{fig:main-figure}):
(1) chunking from text into logical propositions; 
(2) anchoring certain phrases as cues; and finally
(3) integrating these cues through a verbalized explanation.
%
%
Further, we introduce \reward, an intermediate reward strategy for RL that approximates reasoning supervision by counting occurrences of relation-relevant phrases against a credit dictionary. To build this credit dictionary, we extract relational cues from explanations of true-positive samples generated by vanilla LLMs. We show that these cues provide a more stable reward signal, most likely because they are derived from the model’s own training distribution rather than human annotations.


%
%
By providing fine-grained supervision for the reasoning process, this intermediate strategy complements reinforcement learning with verifiable rewards (RLVR)~\citep{Guo_2025}, where rewards based only on the final answer or reasoning format offer limited supervision for the reasoning content without using LLM-as-a-Judge.

In summary, our contributions are fourfold:
\begin{enumerate}[leftmargin=10pt, itemsep=0pt, topsep=1pt]
    \item We propose a novel RE framework loosely inspired by structured cognition. This bottom-up design mitigates LLM attention misalignment when handling complex sentences.
   \item We design \reward, an RL intermediate reward strategy that provides fine-grained supervision for LLM reasoning by matching reasoning cues against a self-generated phrase dictionary.
   \item We introduce a dual evaluation protocol that evaluates both RE performance as well as the quality of the accompanying explanations.  \ours outperforms strong RE baselines by up to +47.1\% F1 (relative); RL with \reward further boosts performance by up to +187.1\% (relative) over the baseline on realistic one-shot TACRED and NYT29,
   while improving human-rated explanation quality by 24.72\% and 54.24\% (relative).
   \item Our pilot experiment identifies a main failure of LLMs on RE as a mismatch between the inferred relations and RE annotations granularity (e.g., failing to distinguish geographic 
   scales in \textsc{org:city\_of\_headquarters}). Human analysis shows that adding \reward reward improves explanation alignment with RE labels in 37.5\% of cases. Models trained with \reward tend to include relational cues aligned with gold labels (e.g., {\em enroll} and {\em attend} for \textsc{per:schools\_attended}), while models without it generate vague terms like \emph{associated}.
\end{enumerate}

\section{Related Work}\label{sec:related-work}

\header{Explainable Relation Extraction} 
Relation extraction is applied in high-stakes domains~\citep{adadi2018peeking,goodman2017european}, where explainability is critical.
Traditional RE models, including feature-based methods, neural networks, and small language models, provide explainability through attention weights~\citep{zhou2016attention}, feature importance~\citep{kambhatla2004combining}, post-hoc analysis~\citep{shahbazi2020relationextractionexplanation}, or rules~\citep{vacareanu2024bestworldspliablegeneralizable,Tang_2023}, but offer limited explainability due to the lack of language-based explanations.

\header{LLM Reasoning} 
Explicit reasoning in LLMs enhances explainability through human-readable reasoning traces~\citep{wei2022chain,chu2025domaino1sguidingllmreasoning}. 
For RE, recent work leverages LLMs via few-shot prompting~\citep{wan2023gpt,ma2023chain,wadhwa2024revisitingrelationextractionera,xu2023unleashpowerlargelanguage}, structured multi-step reasoning~\citep{li2023revisitinglargelanguagemodels}, and instruction tuning~\citep{ouyang2022training,qi2024adelie, jiao2023instructextractinstructiontuning}, but typically relies on handcrafted relation labels and descriptions. 
Recently, reinforcement learning with verifiable rewards has improved both accuracy and explainability~\citep{he2025rethinkingreasoningqualitylarge}. However, explanation-oriented rewards remain unexplored. Existing methods rely either on simple format-based signals~\citep{dai2025r1recrossdomainrelationextraction, wen2025lightr1curriculumsftdpo,xin2025surrogatesignalsformatlength} or costly LLM-as-a-judge approaches~\citep{saha2025learningplanreason,huang2025thinkjlearningthinkgenerative,zhu2026deeprubric,yang2026rubricdropoutsimpleway}.
In this work, we design a cognitively structured reasoning framework for RE without such annotations, and propose an RL self-rewarding strategy that provides low-cost yet fine-grained supervision.


\header{Hints from Cognitive Psychology}
Existing work shows that cognitive psychology provides insights for LLMs and evidences their cognitive capabilities~\citep{yax2024studying, niu2024large}. 
Cognitive psychology has extensively studied how humans process information.
The Construction-Integration model describes comprehension in four steps: forming concepts, elaborating, inferring new propositions, and integrating them into a representation~\citep{kintsch1988role}. 
Several studies show: \textit{chunking} reduces cognitive load~\citep{miller1956magical}, \textit{phrase anchors} guide attention~\citep{kintsch1978toward}, and cognitive monitoring improves strategy adjustment~\citep{flavell1979metacognition}. Inspired by these, we frame RE as a three-step framework, including \textit{chunking}, \textit{anchoring}, and \textit{integrative reasoning}.


\section{Method}\label{sec:method}

Our pilot experiments (\S~\ref{sec:error_analysis}) reveal that LLMs always conduct token-level matching between two sentences and overlook the semantics that truly convey the relation. To address this gap, we design a framework loosely inspired by cognitive science.

\subsection{Cognitive-Structured RE}\label{sec:cog-re}

As shown in Figure~\ref{fig:main-figure} (Part 2), our Cognitive-Structured RE (\ours) framework formulates RE reasoning into three steps.
First, \textit{Proposition Chunking}, where the LLM summarizes each sentence into a relational proposition. 
This step ensures that the LLMs' reasoning starts with compressed propositions rather than long token sequences.
Next, \textit{Phrase Anchoring}, where the LLMs anchor relational phrases in the input sentences and propositions, which grounds the LLMs’ relation-matching reasoning in the original sentence and the extracted propositions. 
The final step is \textit{Integrative Reasoning}.
The LLMs are prompted to integrate propositions and relational phrases into a coherent reasoning chain.
Formally, let the LLM $\mathcal{M}$ be parameterized by $\theta$. Given an input pair $x=(s_1,s_2)$, where $s_1$ is the support sentence and $s_2$ is the test sentence, the model generates an explanation $\hat{z}$ followed by a final label $\hat{y}\in\{\texttt{Yes}, \texttt{No}\}$. The label indicates whether the two sentences express the same relation.
The generation process can be formulated as $(\hat{z},\hat{y}) \sim \mathcal{M}_\theta(\cdot \mid s_1,s_2)$.

\subsection{Reinforcement Learning with \reward}

In RL training, improving the quality of reasoning without a fine-grained reward is challenging, while format-based rewards that ignore reasoning content 
provide only weak guidance. 
To overcome these limitations, we propose an efficient and effective intermediate reward strategy, namely \reward, and integrate it with the accuracy reward to incentivize LLMs' reasoning in RE tasks.

\header{Reward Function Design} 
We design a reward function with two complementary components. 
The first part is the \reward reward, which evaluates the occurrences of relational phrases in the LLM's explanation based on the predefined credit dictionary.
The second part is the accuracy reward, which evaluates the correctness of the predicted results. 
The final reward is formulated as: 
\begin{equation}
\mathcal{R} = \mathbb{I}\!\left(\mathcal{R}_{\text{Acc}} \ge 1\right)\cdot \mathcal{R}_{\text{Hit@Dict}} + \mathcal{R}_{\text{Acc}}
\end{equation}
Here, $\mathcal{R}_{\text{Acc}}$ denotes the accuracy reward, while
$\mathcal{R}_{\text{Hit@Dict}}$ judges the explanation quality.
The indicator function $\mathbb{I}(\cdot)$ is used to gate the explanation reward,
i.e., $\mathcal{R}_{\text{Hit@Dict}}$ is triggered only for correct predictions.
This ensures that the LLM is incentivized to generate correct predictions and explanations aligned well with relational knowledge.

\begin{algorithm*}[t!]
\small
\setlength{\baselineskip}{0.97\baselineskip}
\caption{Building the Relational Phrase Dictionary}
\label{alg:build_keywords_dict}
\begin{algorithmic}[1]
\Require Training set $\mathcal{D}_{\text{train}}=\{(s_1, s_2, r_1, r_2, y)\}$; 
vanilla LLM $\mathcal{M}$; GPT-4o API; 
sample size per label $K \in \{1,\dots,5\}$
\Ensure Phrase dictionary $\mathrm{Dict}: \text{relation\_label} \mapsto \text{phrase list}$
\State $\mathcal{C} \gets \{(s_1, s_2, r_1, r_2, y) \in \mathcal{D}_{\text{train}} \mid r_1 = r_2 \ \wedge\ y=\text{``Yes''}\}$ \CommentBlue{Pairs with identical relation labels}
\State $\mathcal{G} \gets \emptyset$ \CommentBlue{Good cases predicted correctly by the vanilla LLM}
\ForAll{$(s_1, s_2, r_1, r_2, y) \in \mathcal{C}$}
    \State $(\hat{z}, \hat{y}) \gets \mathcal{M}(s_1, s_2)$ \CommentBlue{Vanilla LLM inference: explanation \& label}
    \If{$\hat{y}=\text{``Yes''}$}
        \State $\mathcal{G} \gets \mathcal{G} \cup \{(s_1, s_2, r_1, \hat{z})\}$ \CommentBlue{Keep good cases}
    \EndIf
\EndFor
\State Group $\mathcal{G}$ by relation label: for each $r \in \mathcal{R}$, let $\mathcal{G}_r = \{(s_1, s_2, r, \hat{z}) \in \mathcal{G}\}$
\State $\mathrm{Dict} \gets \emptyset$
\ForAll{$r \in \mathcal{R}$}
    \State $\mathcal{S}_r \gets \text{SampleUpTo}(\mathcal{G}_r, K)$ \CommentBlue{Randomly sample $1\!\sim\!5$ good cases per label}
    \State $\mathrm{prompt}_r \gets \text{BuildPrompt}(r,\ \mathcal{S}_r)$ \CommentBlue{Prompt includes the label and several examples}
    \State $\mathrm{phrases}_r \gets \text{GPT-4o}(\mathrm{prompt}_r)$ \CommentBlue{Generate relational phrases list}
    \State $\mathrm{labelPhrases}_r \gets \text{Tokenize}(\mathrm{label}_r)$ \CommentBlue{Decompose the relation label into phrases}
    \State $\mathrm{Dict}[r] \gets \text{PostProcess}(\mathrm{phrases}_r \cup \mathrm{labelPhrases}_r)$ \CommentBlue{Lowercasing, dedup, stemming/lemmatization, stopword removal}
\EndFor
\State \Return $\mathrm{Dict}$
\end{algorithmic}
\vspace{-2pt}
\end{algorithm*}

\header{\reward Reward}\label{sec:hit_reward}
As shown in Part 2 of Figure~\ref{fig:main-figure}, the \reward reward measures how many relational phrases in model-generated explanation $\hat{z}$ match a pre-built relational phrases dictionary. 
This dictionary is a core component, which collects relation labels appearing in the training dataset, together with their associated relational phrases.

\textit{How can the \reward reward be applied?}
Given an explanation $\hat{z}$ and a relation label $r$, we compute the \reward score as follows.  
Let $\text{Ent}(r)$ denotes the set of entity-related phrases and $\text{Rel}(r)$ the set of relational phrases associated with $r$.  
We define the weighted hit counts as:
\begin{equation}
    \begin{aligned}
        \mathcal{H}_{\text{ent}}(\hat{z}, r) & = \sum_{p \in \text{Ent}(r)} \mathbf{1}[p \in \hat{z}] \\
        \mathcal{H}_{\text{rel}}(\hat{z}, r) & = \sum_{p \in \text{Rel}(r)} \mathbf{1}[p \in \hat{z}]
    \end{aligned}
\end{equation}
where $\mathbf{1}[\cdot]$ is an indicator function that equals $1$ if phrases $p$ appears in $\hat{z}$, and $0$ otherwise.  
For the special case $r=\textsc{no\_relation}$, we set $\mathcal{H}_{\text{ent}} = \mathcal{H}_{\text{rel}}$.
The total weighted hits are given by:
\begin{equation}
    \mathcal{H}(\hat{z}, r) = w_{\text{ent}} \cdot \mathcal{H}_{\text{ent}}(\hat{z}, r) + 
              w_{\text{rel}} \cdot \mathcal{H}_{\text{rel}}(\hat{z}, r)
\end{equation}
with two hyperparameters $w_{\text{ent}}$ and $w_{\text{rel}}$.  
Let $|\hat{z}|$ denote the number of words in $\hat{z}$, normalized by a hyperparameter of $N$. Hyperparameter studies are in Section~\ref{hyperparameter_test}.
The final score is defined as:
\begin{equation}
    \mathcal{S}(\hat{z}, r) = \frac{\mathcal{H}(\hat{z}, r)}{|\hat{z}|/N}
\end{equation}
Finally, the overall \reward reward aggregates the contributions from both sentences $s_1$ and $s_2$, calculated as $\mathcal{R}_{\text{\reward}} = (\mathcal{S}(\hat{z}, r_1) + \mathcal{S}(\hat{z}, r_2)) /2$. 
Here $r_1$ and $r_2$ denote the ground truth relation labels for $s_1$ and $s_2$.
\reward scoring example in Figure~\ref{fig:main-figure}.

\textit{How to construct a dictionary of relational phrases?}
Unlike human-crafted phrases or rules, these phrases are automatically derived from explanations included in the outputs of vanilla LLMs.
In particular, we start by sampling all positive items for which the support and test sentences share the same relation label. Then, the vanilla model answers all the positive items, and we filter the true positive items with the final answer \texttt{Yes}.
For each label that appears in the training dataset, we randomly sample one to five LLM-generated explanations. These relation labels, combined with their associated explanation cases, are input into a LLM,\footnote{We test both Qwen2.5-14B-Instruct and GPT-4o for phrase extraction, obtaining identical results ( Appendix~\ref{appendix:dictionary_robustness_analysis}).} which is prompted to extract the representative relational phrases from these cases.
Additionally, each relation label is decomposed into phrases as part of the relational phrases.
After text post-processing, these relation labels and their associated phrases are added to the relation phrase dictionary.
See Alg.~\ref{alg:build_keywords_dict} and Figure~\ref{fig:main-figure} for details and an example.


\header{Accuracy Reward}
We introduce the accuracy reward $\mathcal{R}_{\text{Acc}}(\hat{y}, y)$ to evaluate the correctness of the predicted label.
In one-shot settings, each test sentence is matched with $K$ supports, with at most one positive and the rest negative, leading to a 1:$K$ imbalance.
Following~\citep{lin2019deepreinforcementlearningimbalanced}, to counter this, we weigh rewards for \textsc{Yes} predictions more heavily, encouraging the model to align with the task’s inherent imbalance. The specific reward values are chosen based on these intuitive observations and are fixed throughout training:
\begin{equation}
\mathcal{R}_{\text{Acc}}(\hat{y}, y) =
\begin{cases}
3.0, & \text{if } \hat{y} = \text{Yes} \land y = \text{Yes} \\
1.0, & \text{if } \hat{y} = \text{No} \land y = \text{No} \\
-3.0, & \text{if } \hat{y} = \text{Yes} \land y = \text{No} \\
-1.0, & \text{if } \hat{y} = \text{No} \land y = \text{Yes} \\
0.0, & \text{otherwise}
\end{cases}
\label{eq:answer_score}
\end{equation}
\header{Training Process}
We optimize \ours with Group Relative Policy Optimization (GRPO)~\citep{shao2024deepseekmathpushinglimitsmathematical}. 
Formally, given a group of $m$ explanation and label pairs $\mathcal{O} = \{(\hat{z}_i, \hat{y}_i) \mid i \in [m]\}$ sampled from \ours for the same input $(s_1, s_2)$, we assign each pair a reward $R(\hat{z}_i, \hat{y}_i)$ using our designed function $\mathcal{R}$.
GRPO encourages relative improvements within a group by normalizing each reward against the group mean. 
Specifically, the \textit{group-relative advantage} of the $i$-th explanation--label pair is defined as:
\begin{equation}
    \mathcal{A}_i =
    \frac{\mathcal{R}(\hat{z}_i,\hat{y}_i) - \frac{1}{m} \sum_{j=1}^m \mathcal{R}(\hat{z}_j,\hat{y}_j)}
         {\text{std}(\{\mathcal{R}(\hat{z}_j,\hat{y}_j) \mid j \in [m]\})},
\end{equation}
where $\text{std}(\cdot)$ is the standard deviation of group rewards.
The overall GRPO objective is optimized to maximize a clipped function with a KL penalty:
\begin{multline}
\mathcal{L}(\theta)
= \mathbb{E}_{(\hat{z}_i,\hat{y}_i)\sim \mathcal{O}}
\Big[
\min\!\Big(
    \rho_i\,\mathcal{A}_i,\; \\
    \text{clip}\!\left(\rho_i, 1-\epsilon, 1+\epsilon\right)\mathcal{A}_i
\Big)
- \beta\,\text{KL}(\theta \,\|\, \theta_{\text{ref}})
\Big].
\end{multline}
$\rho_{i}$ is the importance ratio between updated and old policy probabilities, 
$\epsilon$ controls the clipping range, and $\beta$ weights the penalty for diverging from a reference model $\theta_{\text{ref}}$.

\begin{table*}[h!]
\captionsetup{skip=4pt}
\centering
\setlength{\tabcolsep}{10pt}
\renewcommand{\arraystretch}{1.1}
\begin{adjustbox}{width=0.97\textwidth,center}
\begin{tabular}{@{}p{7cm}|p{0.8cm}p{0.8cm}p{0.8cm}p{0.8cm}p{0.8cm}p{0.8cm} | p{0.8cm}p{0.8cm}p{0.8cm}}
\toprule
\multirow{2}{*}{\textbf{Method}} 
& \multicolumn{3}{c}{\textbf{One-shot TACRED}}
& \multicolumn{3}{c}{\textbf{One-shot NYT29}} & \multicolumn{3}{c}{ \textbf{Avg.}} \\
\cmidrule(lr){2-4} \cmidrule(lr){5-7} \cmidrule(lr){8-10} 
& Prec\% & Recall\% & F1\% & Prec\% & Recall\% & F1\% & Prec\% & Recall\% & F1\% \\ 
\midrule
Semantic Rule Matcher                         & 32.45 & 19.72 & \cellcolor{nblue!30}{24.52}  
                                        & 22.23 & 13.45 & \cellcolor{nblue!10}{16.76} & 27.34 & 16.59 & \cellcolor{nblue!20}{20.64}  \\
SUMASK \emph{(Phi-4)}                                  & 4.44 & 31.71 & 7.78  
                                        & 10.96 & 26.13 & \cellcolor{nblue!10}{15.44} & 7.70 & 28.92 & 11.61  \\
\midrule
\rowcolor{Gainsboro}
\multicolumn{10}{c}{\emph{\textbf{Phi-4}}} \\

Direct Matching                         & 5.43 & 26.83 & 9.03  
                                        & 9.04 & 23.15 & 13.00 & 7.24 & 24.99 & 11.02  \\
Simple Reasoning                        & 5.69 & 58.54 & \cellcolor{nblue!10}{10.38}  
                                        & 8.71 & 30.68 & \cellcolor{nblue!10}{13.57} & 7.20 & 44.61 & \cellcolor{nblue!10}{11.98}  \\
\textbf{\ours (\emph{ours})}            & 22.53 & 50.00 & \cellcolor{nblue!20}{31.06}  
                                        & 12.03 & 30.68 & \cellcolor{nblue!20}{17.28} & 17.28 & 40.34 & \cellcolor{nblue!30}{24.17} \\
\noalign{\kern 2pt}
\cdashline{1-10}[3pt/3pt]
\noalign{\kern 2pt}

\textbf{\ours After RL with Acc}            & 26.90 &47.56 & \cellcolor{nblue!30}{34.36}  
                                        & 45.50 & 37.36 & \cellcolor{nblue!50}{41.02} & 36.20 & 42.46 & \cellcolor{nblue!40}{37.69}  \\


\textbf{\ours After RL with \reward + Acc} & 26.88 & 60.98 & \cellcolor{nblue!40}{37.31}  
                                        & 45.14 & 44.89 & \cellcolor{nblue!60}{45.01} & 36.01 & 52.94 & \cellcolor{nblue!50}{41.16}  \\
\midrule
\rowcolor{Gainsboro}
\multicolumn{10}{c}{\emph{\textbf{Qwen2.5-14B-Instruct}}} \\
Direct Matching                         & 13.33 & 2.44 & 4.12  
                                        & 48.73 & 13.63 & \cellcolor{nblue!16}{21.31} & 31.03 & 8.04 & 12.72  \\
Simple Reasoning                        & 5.67 & 34.15 & \cellcolor{nblue!10}{9.72}  
                                        & 11.85 & 34.23 & 17.61  & 8.76 & 34.19 & 13.67  \\
\textbf{\ours (\emph{ours})}            & 29.49 & 28.05 & \cellcolor{nblue!30}{28.75}  
                                        & 20.18 & 31.67 & \cellcolor{nblue!16}{24.65} & 24.84 & 29.86 & \cellcolor{nblue!16}{26.70}  \\
\noalign{\kern 2pt}
\cdashline{1-10}[3pt/3pt]
\noalign{\kern 2pt}

\textbf{\ours After RL with Acc} & 26.83 &40.24 & \cellcolor{nblue!45}{32.20}  
                                        & 26.17 & 29.40 & \cellcolor{nblue!30}{27.69} & 26.50 & 34.82 & \cellcolor{nblue!30}{29.95} \\


\textbf{\ours After RL with \reward + Acc} & 22.08 & 62.20 & \cellcolor{nblue!45}{32.58}  
                                        & 63.34 & 38.78 & \cellcolor{nblue!70}{48.11} & 42.71 & 50.49 & \cellcolor{nblue!60}{40.34} \\
\bottomrule

\end{tabular}
\end{adjustbox}
\caption{Precision (P), recall (R), and F1 on the one-shot TACRED and one-shot NYT29 datasets. 
We split the table into three blocks: baseline methods; prompting methods without RL, including vanilla prompting and our \ours framework; and models trained with RL on top of the same \ours prompting, with the accuracy reward alone or combined with \reward.
We use fixed $N=5$, $w_{\text{ent}}=0.4$, $w_{\text{rel}}=1.0$; hyperparameter experiments in Table~\ref{tab:hyperparameters_test}.
The \bluetext{blue color} highlights F1 scores, with darker (from \shallowbluetext{  } to \bluetext{   }) shades indicating larger values. }
\label{tab:main-results}
\vspace{-10pt}
\end{table*}


\section{Experimental Setup}\label{sec:experiment-setup}

\noindent\textbf{Benchmark.} 
Our work focuses on realistic Few-Shot Relation Extraction (FSRE)~\cite{alam2024realisticfewshotrelationextraction}, which predicts relation labels for test sentences from several support sentences without predefined labels or descriptions. This setting is more challenging than standard FSRE due to minimal supervision (see Appendix~\ref{appendix:realistic_FSRE}). 
We follow this realistic FSRE and conduct experiments on \textit{one-shot TACRED, NYT29, and WIKIDATA}.
The relation labels in the training and testing partitions are out-of-distribution. 
Besides, since traditional RE methods rely on small classifiers, RE benchmarks are built to be extremely large. 
Following previous work~\citep{li2023revisitinglargelanguagemodels}, we randomly sampled 1,000 episodes per partition according to the original label distribution (Statistics in Appendix~\ref{appendix:testing_set_distribution}).

\noindent\textbf{Evaluation.} 
We adopt a dual evaluation protocol containing both automatic and human evaluation.
We use the F1 score as the automatic evaluation metric.
For human evaluation, we rated explanations on a 3-point Likert scale: 2 points for the correctness and conciseness of the summaries, plus 1 point if it aligns with the RE labeling. The detailed evaluation rubric is provided in Appendix~\ref{appendix:human_evaluation_rubric}.
Two annotators with NLP backgrounds independently rated the sampled explanations. The Cohen’s kappa score is 0.693, indicating substantial agreement and that our evaluation rubric is well-defined.

\noindent\textbf{Baselines.} We follow the evaluation method in Realistic FSRE and compare our method with two categories of baselines.
\textbf{\textit{Conventional RE Models}}: Semantic Rule Matcher~\citep{vacareanu2024bestworldspliablegeneralizable}, which combines a neural classifier with rules, achieving state-of-the-art results on realistic Few-Shot TACRED and NYT29. 
\textbf{\textit{Prompting RE baselines}}:
(i) SUMASK~\citep{li2023revisitinglargelanguagemodels} reformulates relation extraction as a question answering task. We implement SUMASK under realistic FSRE constraints (no label) ; see Appendix~\ref{appendix:SUMASK_prompt}.
(ii) Naive prompting: we include \textit{direct-matching} (i.e., only producing \texttt{Yes}/\texttt{No}) and \textit{simple-reasoning} (i.e., reasoning before \texttt{Yes}/\texttt{No}); see Appendix~\ref{appendix:naive_prompt}. 

\noindent\textbf{Implementation Details.}\label{sec:implement_detail}
We sample 20,000 items from the training partition, preserving the distribution of relation labels and maintaining an approximate 1:7 ratio between positive and negative instances (statistics in Appendix~\ref{appendix:training_set_distribution}).
We implement our method with Qwen2.5-14B-Instruct and Phi-4 using fixed reward hyperparameters,
$N=5$, $w_{\text{ent}}=0.4$, and $w_{\text{rel}}=1.0$ in the main experiments reported in Table~\ref{tab:main-results}.
To assess robustness, we further evaluate our method under different reward hyperparameter settings,
with $N \in \{3,5,7\}$, $w_{\text{ent}} \in \{0.2,0.4,0.6\}$, and
$w_{\text{rel}} \in \{0.8,1.0,1.2\}$, and report the results in Table~\ref{tab:hyperparameters_test}.
%
%
We optimize the model using Verl~\citep{Sheng_2025} with an actor learning rate of $1 \times 10^{-6}$, KL regularization (coefficient 0.01). Training is conducted on 4 NVIDIA H100 80 GB GPUs. A complete run on the 14B--15B model takes~20 GPU-hours.

\section{Experimental Results}\label{sec:experiment-results}

\subsection{Main Results}

Table~\ref{tab:main-results} reports the main results under different reasoning designs and RL settings, using either accuracy reward alone or combined with \reward.

\noindent\textbf{\ours improves accuracy with balanced precision and recall.}
Our \ours consistently outperforms all baselines with a better precision–recall trade-off.
In contrast, the Semantic Rule Matcher (the previous SOTA), based on rules and a small language model, yields high precision but lower recall.
Prompting-based LLM baselines rely solely on LLMs' generalization ability, leading to high recall but lower precision. 
\ours combines both perspectives: it anchors reasoning with 
phrases while leveraging LLMs' generalization through summarization and integrative reasoning. 

\noindent\textbf{\reward reward improves task accuracy.}
Table~\ref{tab:main-results} shows that Qwen2.5-14B-Instruct, trained only with an accuracy reward, surpasses its non-trained version by +3.45\% and +3.04\%, while the trained Phi-4 leads to +3.30\% and +23.74\% improvements.
\reward + Acc reward further boosts accuracy across models, which outperforms accuracy-only training and achieves the best results.
In particular, Qwen2.5-14B-Instruct achieves F1=48.11\% with \reward + Acc, a 73.74\% relative gain over the accuracy-only version. 
We also evaluate the quality of the explanations (see \S~\ref{sec:error_analysis}).

\begin{figure*}[t!]
    \captionsetup{skip=4pt}
    \centering
    \begin{subfigure}{0.33\textwidth}
        \includegraphics[width=\linewidth]{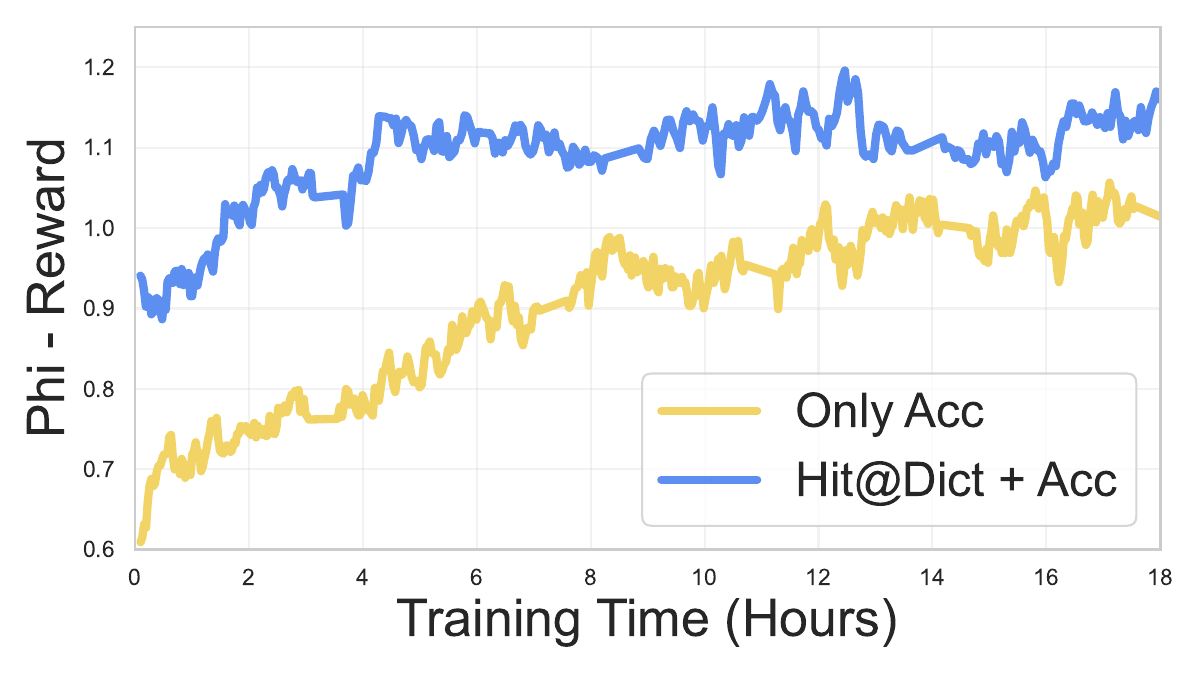}
        \caption{Phi-Reward}
    \end{subfigure}
    \hspace{-0.3cm}
    \begin{subfigure}{0.33\textwidth}
        \includegraphics[width=\linewidth]{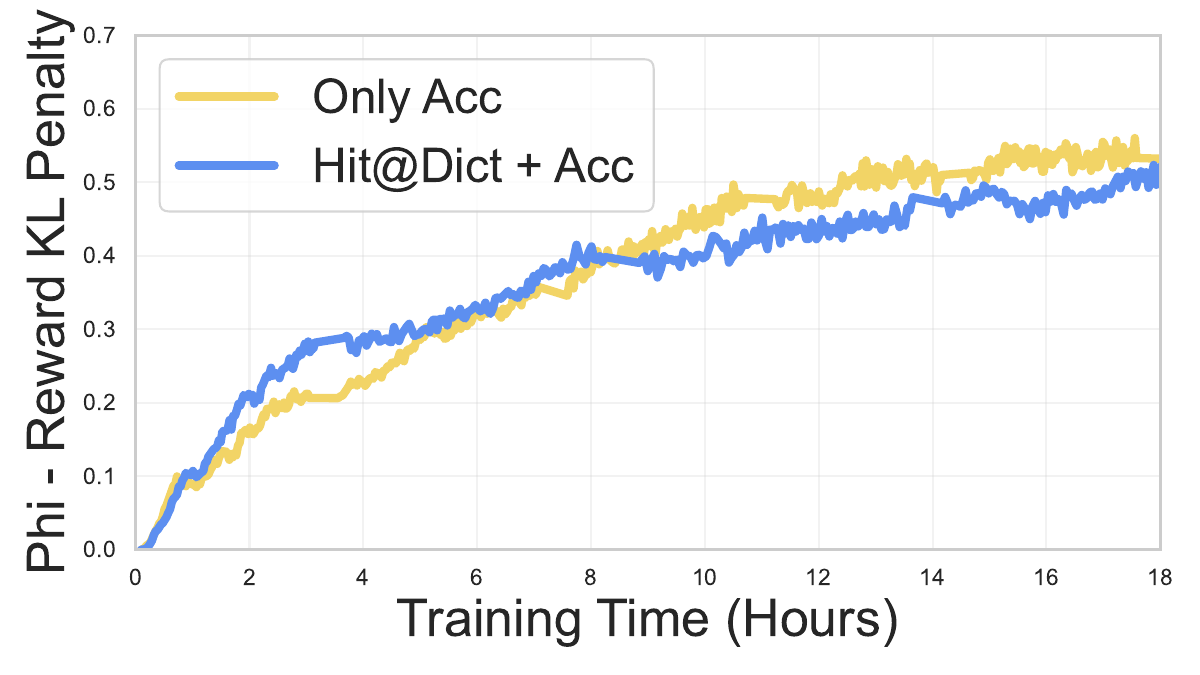}
        \caption{Phi-KL}
    \end{subfigure}
    \hspace{-0.3cm}
    \begin{subfigure}{0.33\textwidth}
        \includegraphics[width=\linewidth]{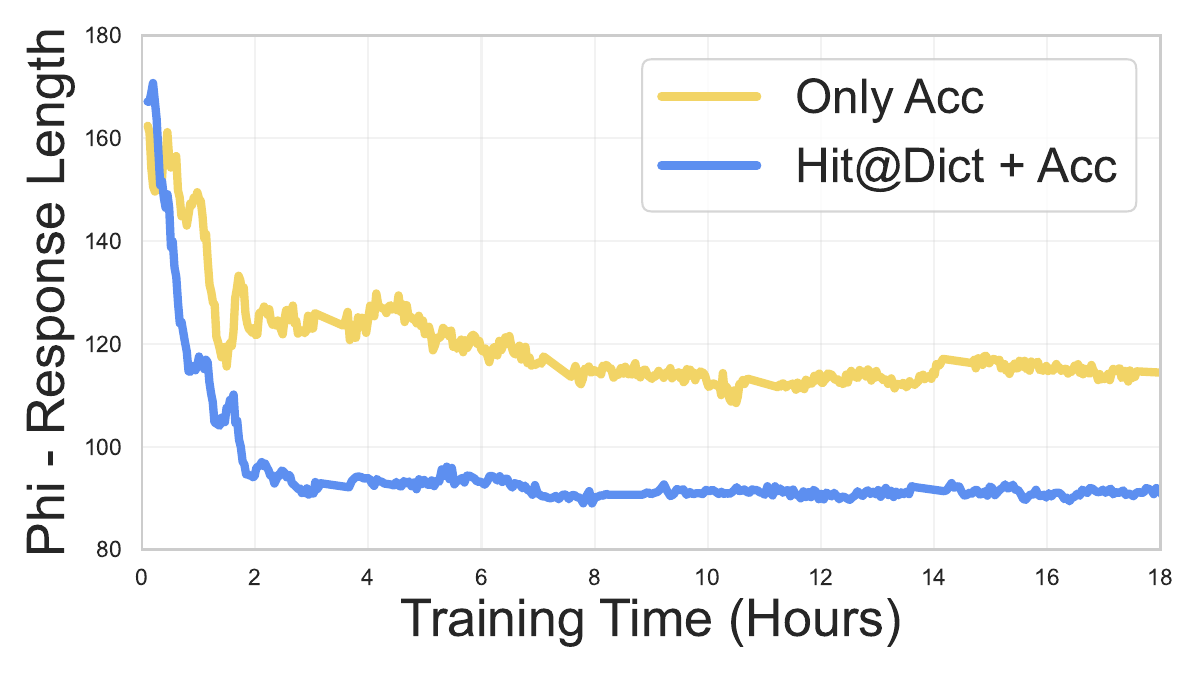}
        \caption{Phi-Response Length}
    \end{subfigure}
    \hspace{-0.3cm}
    \begin{subfigure}{0.33\textwidth}
        \includegraphics[width=\linewidth]{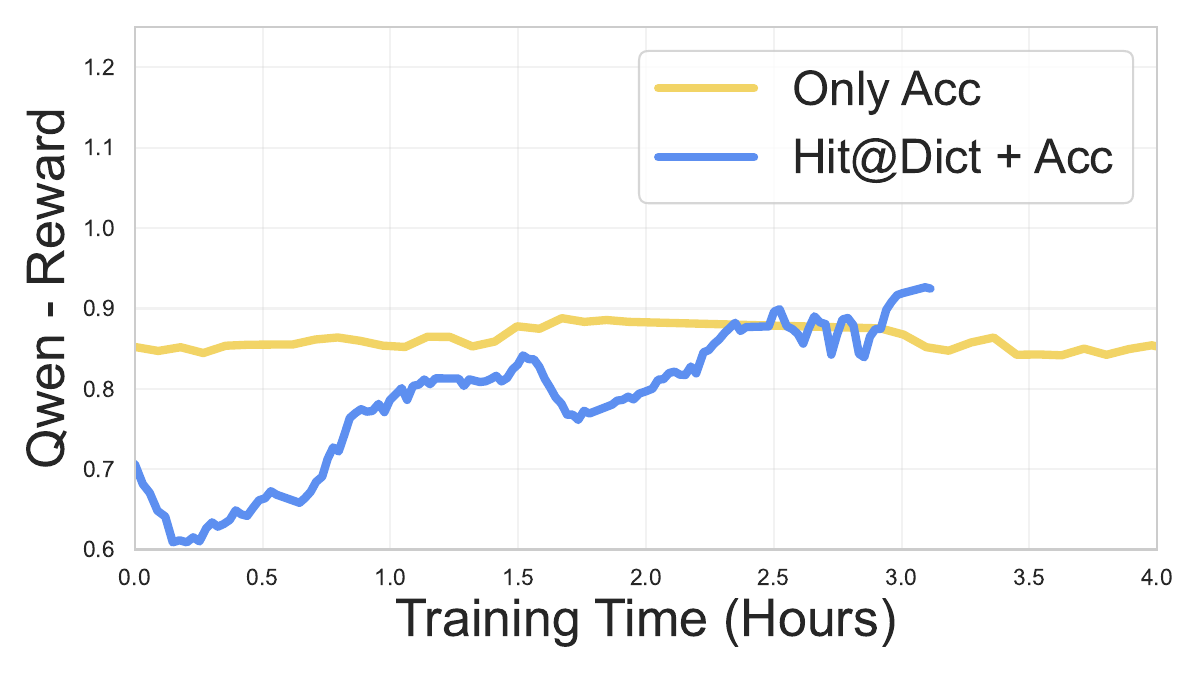}
        \caption{Qwen-Reward}
    \end{subfigure}
    \hspace{-0.3cm}
    \begin{subfigure}{0.33\textwidth}
        \includegraphics[width=\linewidth]{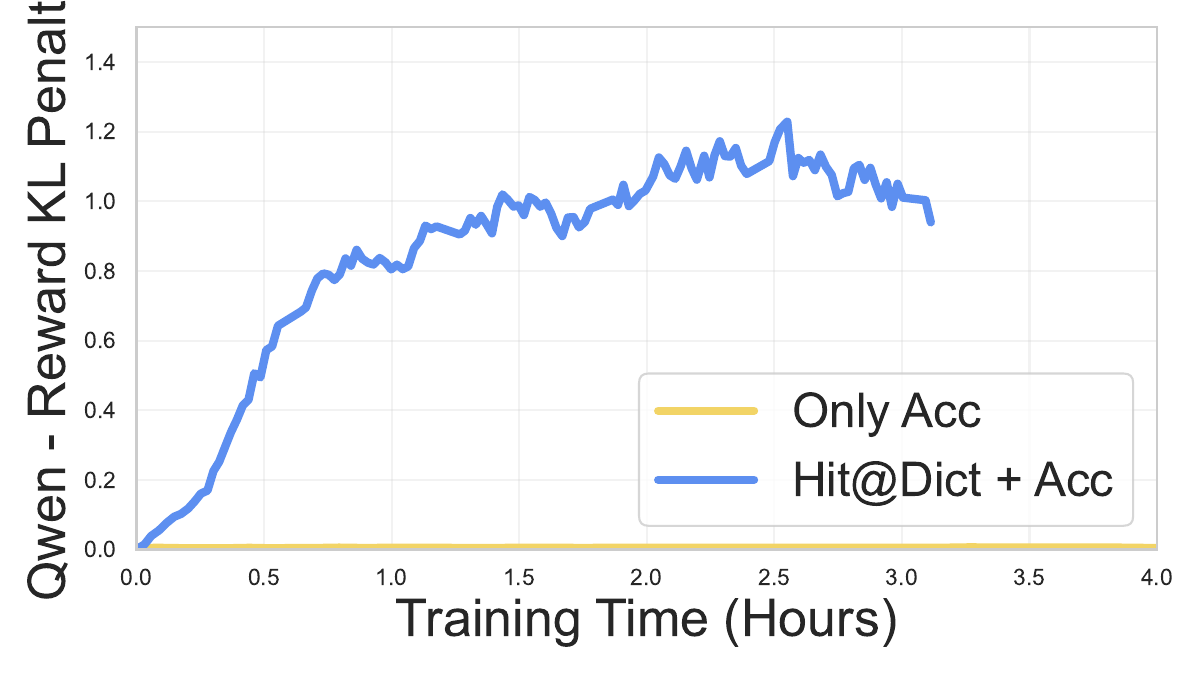}
        \caption{Qwen-KL}
    \end{subfigure}
    \hspace{-0.3cm}
    \begin{subfigure}{0.33\textwidth}
        \includegraphics[width=\linewidth]{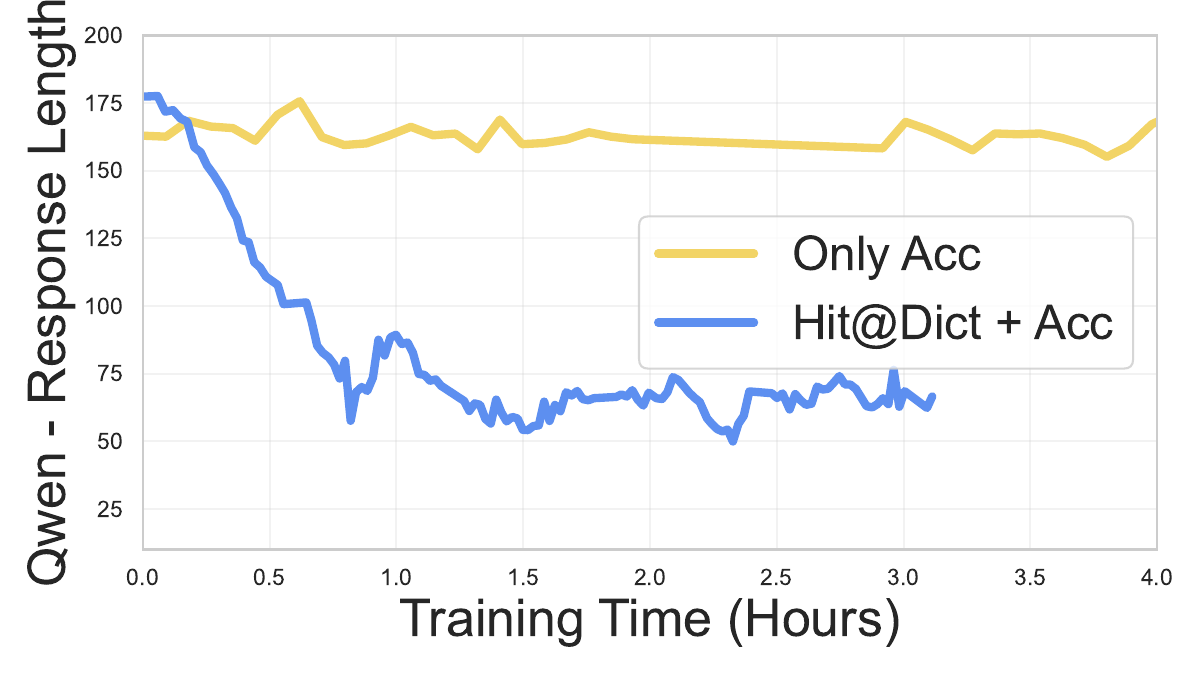}
        \caption{Qwen-Response Length}
    \end{subfigure}
    \caption{Training dynamics on the one-shot NYT29 dataset for Phi-4 and Qwen2.5-14B-Instruct. The X-axes show training time(Hours). The Y-axes show reward, KL penalty, and response length. We compare reinforcement learning with accuracy reward \yellowtext{\textit{Only Acc}} and with the combined \reward reward \bluetext{\textit{\reward+Acc}}. We use fixed hyperparameters with $N=5$, $w_{\text{ent}}=0.4$, and $w_{\text{rel}}=1.0$.}
    \label{fig:rl_training_process}
    \vspace{-8pt}
\end{figure*}

\subsection{Behavior of \reward Reward}
We analyze the impact of the \reward and accuracy rewards during the RL training process.
Figure~\ref{fig:rl_training_process} shows the RL training process of both Phi-4 and Qwen2.5-14B-Instruct on the one-shot NYT29. 

\noindent\textbf{\reward accelerates the convergence of training.} 
In Figure~\ref{fig:rl_training_process}(a), \textit{\reward+Acc} curve rapidly increases to above 1.1 within 5 hours and stabilizes between 1.1--1.2 after 10 hours; in contrast, the \textit{Only Acc}  curve climbs slowly from 0.6 and stabilizes around 1.0 after 13 hours.


\noindent\textbf{\reward provides more stable training.}
As shown in Figure~\ref{fig:rl_training_process} (d) and (e), Qwen2.5-14B-Instruct, trained only with the accuracy reward, exhibits stagnant reward values and an extremely small reward–KL penalty, indicating the policy remains almost unchanged from its initial state. In contrast, Qwen2.5-14B-Instruct, trained with the \reward reward, shows steady growth. 

\noindent\textbf{\reward encourages more concise explanations.}
As Figure~\ref{fig:rl_training_process} (c) and (f) show, with the \reward reward, the response length compresses to 75--90 tokens. Combined with our analysis of the explanations, we found that in most cases, the models produce more concise explanations. It enhances reasoning efficiency with concise outputs. 


\begin{table}[!t]
\captionsetup{skip=4pt}
\centering
\setlength{\tabcolsep}{10pt}
\renewcommand\arraystretch{1.1}
\begin{adjustbox}{width=1\columnwidth,center}

\begin{tabular}{p{2.5cm} | p{0.6cm}p{0.6cm}p{0.6cm}p{0.6cm}p{0.6cm}p{0.6cm}}
\toprule
\multirow{2}{*}{\textbf{Method}} 
& \multicolumn{3}{c}{One-shot TACRED}
& \multicolumn{3}{c}{One-shot NYT29} \\
\cmidrule(lr){2-4} \cmidrule(lr){5-7}
& Prec\% & Recall\% & F1\% & Prec\% & Recall\% & F1\% \\
\midrule
\rowcolor{Gainsboro} \ours        & 22.53 & 50.00 & \cellcolor{nblue!40}{31.06}  
                                              & 12.03 & 30.68 & \cellcolor{nblue!40}{17.28}  \\
- \textit{w/o chunking}              & 7.05 & 57.68 & 12.56  
                                               & 8.20 & 27.41 & 12.63  \\
- \textit{w/o phrases}             & 16.10 & 52.44 & \cellcolor{nblue!15}{24.64}  
                                              & 10.39 & 31.39 & \cellcolor{nblue!15}{15.62}  \\
- \textit{w/o reasoning}           & 5.94 & 28.05 & 9.81  
                                              & 9.91 & 24.57 & 14.12  \\

\bottomrule
\end{tabular}
\end{adjustbox}
\caption{Ablation studies on one-shot TACRED and NYT29 with Phi-4.
Darker \bluetext{shades} indicate higher F1.}\label{tab:ablation-results}
\vspace{-2pt}
\end{table}

\subsection{Ablation Study}\label{sec:ablation_experiment}

We analyze the effectiveness of each component in \ours by removing one step at a time:%
(1) \textit{w/o chunking}: removes chunking;
(2) \textit{w/o phrases}: removes phrase anchoring;
(3) \textit{w/o reasoning}: removes reasoning.
Results are shown in Table~\ref{tab:ablation-results}, from which we draw three observations.
\textit{First}, all three steps contribute to the final performance. 
Removing any step from our framework causes a clear performance drop, ranging from –1.66\% to –21.25\%.
\textit{Second}, chunking and phrase anchoring mainly improve precision. Removing either lowers precision on both datasets (–15.48\% and –3.83\% for chunking; –6.43\% and –1.64\% for phrase anchoring), while recall is far less affected and even increases on TACRED (+7.68\% and +2.44\%). This indicates that both steps suppress spurious matches rather than retrieve additional ones.
\textit{Third}, reasoning improves both precision and recall, with a larger impact on recall. 
Removing it substantially reduces recall (–21.95\% on TACRED; –6.11\% on NYT29) alongside precision (–16.59\% and –2.12\%).

\subsection{Out-of-Distribution Generalization}
\begin{table}[t!]
\captionsetup{skip=4pt}
\centering
\setlength{\tabcolsep}{10pt}
\begin{adjustbox}{width=0.97\columnwidth,center}
\begin{tabular}{@{}p{7cm} | p{0.8cm}p{0.8cm}p{0.8cm}}
\toprule
\multirow{2}{*}{\textbf{Method}} 
& \multicolumn{3}{c}{\textbf{One-shot WIKIDATA}}\\
\cmidrule(lr){2-4}
& Prec\% & Recall\% & F1\%\\
\midrule
Semantic Rule Matcher   & 6.54 & 6.08 & 6.31 \\
SUMASK \emph{(Phi-4)}   & 1.48 & 27.45 & 2.80 \\
\midrule
Direct Matching     & 1.13 & 49.02 & 2.22 \\
Simple Reasoning    & 1.99 & 56.86 & 3.85\\
\textbf{\ours (\emph{ours})}  & 5.08 & 52.94 & \cellcolor{nblue!10}{9.28} \\

\noalign{\kern 2pt}
\cdashline{1-4}[3pt/3pt]
\noalign{\kern 2pt}

\textbf{\ours After RL with Acc}            & 14.29  & 31.37 & \cellcolor{nblue!20}{19.63} \\
\textbf{\ours After RL with \reward + Acc}  & 37.84  & 27.45 & \cellcolor{nblue!50}{31.81} \\
\bottomrule

\end{tabular}
\end{adjustbox}
\caption{
Out-of-domain evaluation for Phi-4 trained on the one-shot NYT29 (same as Table~\ref{tab:main-results}) on the one-shot WIKIDATA.
Darker \bluetext{shades} indicate higher F1.}
\label{tab:OOD}
\end{table}

These Phi-4 checkpoints trained on the one-shot NYT29 are further evaluated on the one-shot WIKIDATA.
Results (Table~\ref{tab:OOD}) show that \ours consistently outperforms baseline approaches in F1 score.
RL training with the accuracy-only reward improves F1 to 19.63\% on Phi-4.
Further optimization with our \reward + Acc yields substantial additional gains, improving F1 by +12.18\% on Phi-4. In contrast, when transferring from NYT29 to WIKIDATA, model performance with other methods drops substantially.
For example, Phi-4 exhibits F1 drops of 12.64\% with SUMASK, 10.78\% with direct matching, and 9.72\% with simple reasoning.
We also conduct a qualitative analysis of explanation quality on WIKIDATA, showing that \ours after RL with \reward + Acc generates more concise explanations that align better even with unseen RE relation labels.
These findings demonstrate that our method generalizes robustly to datasets with unseen relation labels.


\subsection{Error Analysis of LLM on RE-reasoning}\label{sec:error_analysis}

\header{Simple reasoning strategy}
We analyze the explanations of vanilla LLMs using a random reasoning strategy. For this stage, we select Qwen2.5-14B-Instruct and GPT-4o. 
From their explanations, we identify two common failure patterns.
\textit{First}, failing to focus on semantics that truly convey the relation.
When matching two sentences, LLMs frequently focus on irrelevant tokens in the second sentence, aiming to align with the relation conveyed in the first.
We provide a simplified example in Table~\ref{tab:error_pattern_1}. 
In this case, LLMs incorrectly focus on two names in the second sentence in order to mimic the relation of \textsc{per:alternate\_name} in the first sentence. 
\textit{Second}, failing to align with the granularity of the RE human-annotation schema. Without the human-crafted descriptions of relation labels, LLMs struggle to distinguish between similar human-defined relations, e.g., \textsc{org:country\_of\_headquarters} and \textsc{org:city\_of\_headquarters}.  

\header{\ours}
We evaluate the quality of LLM-generated explanations at three stages: (1) vanilla LLMs with the \ours, and after RL training, (2) with only an accuracy reward, and (3) with both \reward and accuracy reward. We validate Qwen2.5-14B-Instruct and Phi-4 on two datasets. For each LLM–dataset–stage combination, we sample 40 explanations, with 10 explanations per category (TP, TN, FP, FN). 
Results (Appendix~\ref{appendix:human_evaluation_results}) show that human evaluation scores improve by 54.24\% (relative). Compared with vanilla and accuracy-only, \reward combined with accuracy reward enables more concise summaries and better alignment with human annotations. For example, in the Phi–TACRED setting, among the 40 analyzed cases, the model trained with the \reward + Acc produced more concise summaries in 8 cases and exhibited better alignment with human labeling in 15 cases. Specifically, the trained model tends to include relational phrases closely aligned with gold labels in their explanation (e.g., {\em enroll}, {\em attend}, and {\em university} for the relation \textsc{per:schools\_attended}), while the untrained model generates vague terms such as {\em associated}. See some case comparisons in Appendix~\ref{appendix:cases_comparison_phi}.


\subsection{Hyperparameter Sensitivity Analysis}\label{hyperparameter_test}
\begin{table}[!t]
\captionsetup{skip=4pt}
\centering
\setlength{\tabcolsep}{10pt}
\renewcommand\arraystretch{0.96}
\begin{adjustbox}{width=0.95\columnwidth,center}

\begin{tabular}{p{0.7cm} | p{0.3cm}c c p{1cm}p{1cm}p{1cm}}
\toprule
\multirow{2}{*}{No.} 
& \multicolumn{3}{c}{Hyperparameters}
& \multicolumn{3}{c}{One-shot NYT29} \\
\cmidrule(lr){2-4} \cmidrule(lr){5-7}
& $N$&$w_{\text{ent}}$&$w_{\text{rel}}$& Prec\% & Recall\% & \textit{ }F1\% \\
\midrule
1        & \cellcolor{nyellow}{5} & 0.4 & 1.0 & 63.34 & 38.78 & 48.11  \\
2        & \cellcolor{nyellow}{3} & 0.4 & 1.0 & 43.75 & 56.68 & 49.38  \\
3        & \cellcolor{nyellow}{7} & 0.4 & 1.0 & 44.06 & 52.13 & 47.76  \\
\noalign{\kern 2pt}
\cdashline{1-7}[3pt/3pt]
\noalign{\kern 2pt}
4        & 5 &  \cellcolor{nyellow}{0.2} & 1.0 & 45.73 & 51.70 & 48.53  \\
5        & 5 &  \cellcolor{nyellow}{0.6} & 1.0 & 45.61 & 48.72 & 47.12  \\
\noalign{\kern 2pt}
\cdashline{1-7}[3pt/3pt]
\noalign{\kern 2pt}
6        & 5 & 0.4 & \cellcolor{nyellow}{0.8} & 47.34 & 48.01 & 47.67  \\
7        & 5 & 0.4 & \cellcolor{nyellow}{1.2} & 35.05 & 72.58 & 47.27  \\
\bottomrule
\end{tabular}
\end{adjustbox}
\caption{Hyperparameters sensitivity on one-shot NYT29 with Qwen2.5-14B-Instruct. 
We test seven configurations by \yellowtext{varying} $w_{\text{ent}}$, $w_{\text{rel}}$, $N$ individually.}\label{tab:hyperparameters_test}
\vspace{-2pt}
\end{table}

We analyze the sensitivity of our method to its hyperparameters, including the entity weight $w_{\text{ent}}$, the relation weight $w_{\text{rel}}$, and the normalization factor $N$. 
We vary $N \in \{3, 5, 7\}$, $w_{\text{ent}}\in \{0.2, 0.4, 0.6\}$, and $w_{\text{rel}}\in \{0.8, 1.0, 1.2\}$.
To ensure a fair comparison, all hyperparameter experiments are conducted using Qwen2.5-14B-Instruct on the same NYT29 dataset.
Results under all seven configurations are reported in Table~\ref{tab:hyperparameters_test}.
We find that the model’s performance remains stable across these seven hyperparameter configurations.
The F1 score remains largely consistent at $47.98 \pm 0.78$.
This small variance indicates that our method is robust to changes in hyperparameter configurations.

\section{Conclusion}\label{sec:conclusion}

In this work, we introduced \textit{\ours}, a RE reasoning framework loosely inspired by cognitive science, which decomposes RE into three human-like text-processing steps. To incentivize LLM reasoning to align with RE labeling schemas via RL, we proposed \reward, a lightweight reward strategy that derives reasoning supervision from the model’s own correct predictions. Experiments and human evaluations on one-shot TACRED and NYT29 demonstrate that our framework achieves enhanced accuracy and explanation quality. Human analysis confirms that our reward design encourages models to generate more concise and label-aligned reasoning. 
Future work will extend \reward to more reasoning-intensive tasks.

\newpage
\section*{Limitations}
This work does not investigate the internal causal mechanisms of LLMs, but instead focuses on improving their textual reasoning process. In particular, we do not analyze internal model representations or design methods that directly improve reasoning through representation-level interventions. Mechanistic interpretability approaches, such as sparse autoencoders, may provide useful tools for such analysis. Our work instead focuses on improving explainability through textual reasoning, leaving mechanistic interpretability for future work.

\section*{Ethical Considerations}
This work does not involve human subjects research beyond standard human evaluations. All datasets, and pretrained models used in this work are publicly available and used in accordance with their intended research use and applicable licenses. We use public relation extraction benchmarks and do not collect new personal data. We provide dataset descriptions, evaluation settings, and data statistics in Section 4 and Appendix. AI-assistants were only used to polish the writing, such as improving grammar, readability. All technical ideas, experiments, analysis, and final verification were conducted by the authors.




\section*{Acknowledgments}
Mihai Surdeanu declares a financial interest in \href{https://lum.ai}{lum.ai}. This interest has been properly disclosed to the University of Arizona Institutional Review Committee and is managed in accordance with its conflict of interest policies.

\bibliography{main}



\newpage
\clearpage
\appendix
\section{Appendix}

\section*{Content of Appendix}
\begin{itemize}[noitemsep, topsep=2pt, leftmargin=36pt]
    \item[\ref{appendix:realistic_FSRE}] Realistic Few-Shot Relation Extraction
    \item[\ref{appendix:human_evaluation_rubric}] Human Evaluation Rubric
    \item [\ref{appendix:human_evaluation_results}] Human Evaluation Results
    \item [\ref{appendix:dictionary_robustness_analysis}] Dictionary Robustness Analysis
    \item [\ref{appendix:training_set_distribution}] Statistics of the Sampled Training Set
    \item [\ref{appendix:testing_set_distribution}] Statistics of the Sampled Testing Set
    \item [\ref{appendix:prompt_CogRE}] Prompt for \ours framework
    \item [\ref{appendix:Prompt_for_Relation_Keyword_Extraction}] Prompt for Relation Phrases Extraction
    \item [\ref{appendix:naive_prompt}] Prompt for Baselines
    \begin{itemize}[noitemsep, topsep=2pt]
        \item[\ref{appendix:SUMASK_prompt}] Prompts for SUMASK
        \item[\ref{appendix:Direct-Answer-Prompt}] Direct-Answer Prompt
        \item[\ref{appendix:Simple-Reasoning-Prompt}] Simple-Reasoning Prompt
    \end{itemize}
    \item [\ref{appendix:prompt_for_ablation_experiments}] Prompt for Ablation experiments
    \begin{itemize}[noitemsep, topsep=2pt]
        \item[\ref{appendix:prompt_wo_chunking}] COGRE- w/o chunking Prompt
        \item[\ref{appendix:prompt_wo_reasoning}] COGRE- w/o reasoning Prompt
        \item[\ref{appendix:prompt_wo_keywords}] COGRE- w/o phrases Prompt
    \end{itemize}
    \item [\ref{appendix:inference_results_cross_models}] Inference Results across Model Families and Sizes on realistic One-shot RE task
    \item [\ref{appendix:cases_comparison_phi}] Cases Comparison of Phi-4 on TACRED
    \item [\ref{appendix:keywords_comparison}] Comparison of Extracted Dictionaries by Open-source and Closed-source LLMs
\end{itemize}


\subsection{Realistic Few-Shot Relation Extraction}\label{appendix:realistic_FSRE}
The Relation Extraction (RE) task studied in this paper is the realistic Few-Shot Relation Extraction (FSRE) task~\cite{alam2024realisticfewshotrelationextraction}. This task is fundamentally different from the standard FSRE setting; it is harder and more realistic.  In realistic FSRE, the model infers a relation by comparing a test sentence against several support sentences {\em without predefined labels or label descriptions}. This reflects real-world RE where many relations lack manual descriptions, sentences are unlabeled, and response time is critical.

\subsubsection{Framework of realistic few-shot RE setting}
\begin{figure*}[h]
    \centering
    \includegraphics[width=0.8\linewidth]{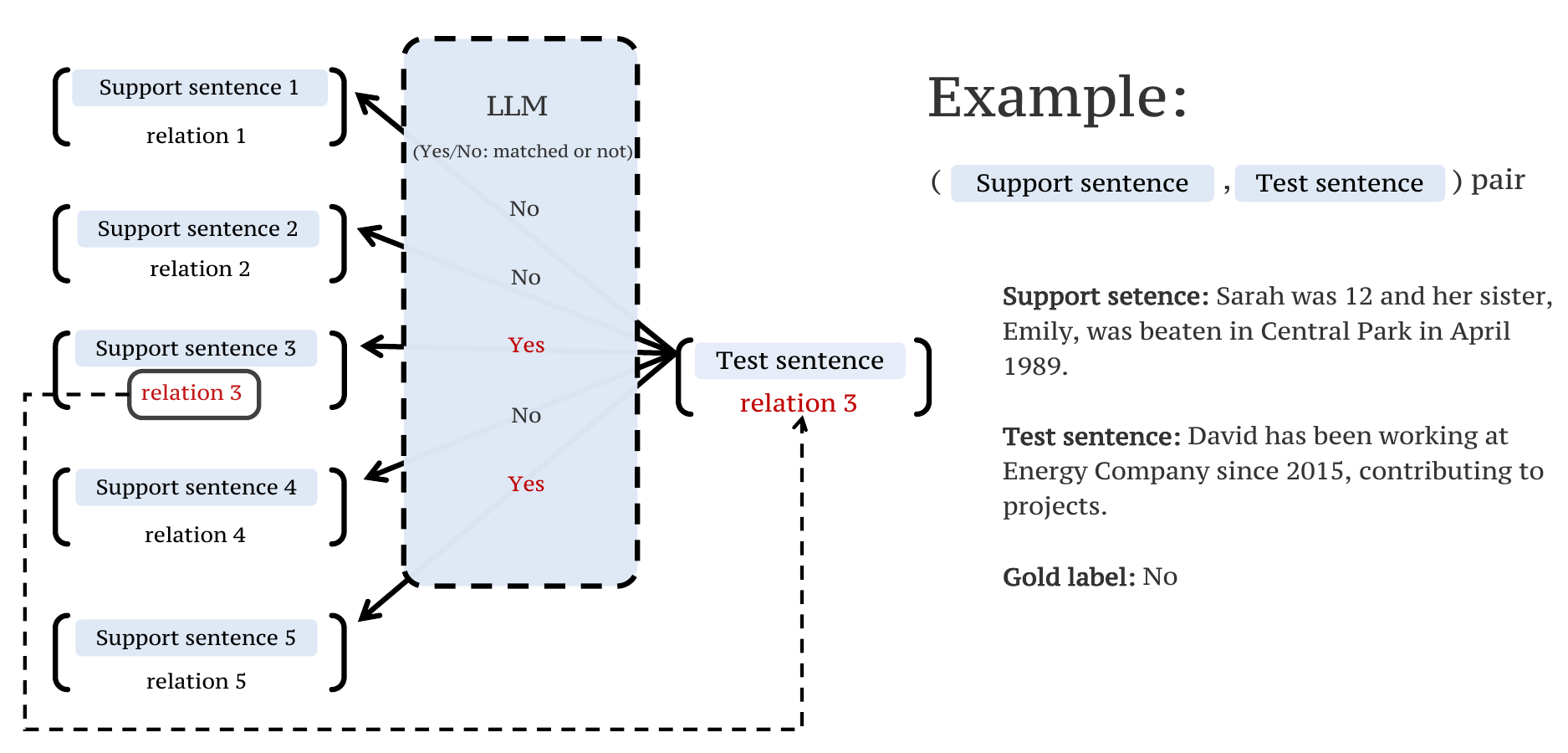}
    \caption{Illustration of the realistic few-shot relation extraction setting.
Pairwise support sentence and test sentence matching results are aggregated to produce a multi-class prediction.}
    \label{fig:task_setting}
\end{figure*}

As shown in Figure~\ref{fig:task_setting} for each relation, we provide one support sentence as an example of that relation.
Given a test sentence, the model compares it against each support sentence individually
and outputs \textit{Yes} or \textit{No}. If all outputs are \textit{No}, the predicted
label is \textit{no\_relation}. If at least one output is \textit{Yes}, the corresponding
relation is selected. The final prediction is then compared with the gold label to
compute the F1 score.

\subsubsection{Characteristics of realistic few-shot RE setting}
This task setting has the following key characteristics:
\textbf{(1) it is a multi-class Relation Extraction task},
\textbf{(2) it follows a realistic few-shot RE setting},
and \textbf{(3) it is substantially more challenging than traditional multi-class RE}.
We provide a detailed explanation below.

\paragraph{(1) Multi-class Relation Extraction.}
The RE task considered in this paper is a multi-class relation classification problem.
Although our method adopts a pairwise matching formulation as an intermediate step,
the overall task remains multi-class RE, and evaluation is based on whether the
predicted relation matches the gold label.

This pairwise formulation is common in few-shot relation extraction.
For example, SUMASK~\cite{li2023revisitinglargelanguagemodels} compares a test sentence with each relation triplet
by generating relation-specific questions and making one-by-one decisions.
Similarly, Semantic Rule Matcher~\cite{vacareanu2024bestworldspliablegeneralizable} performs pairwise
matching between support and test sentences before producing a final multi-class
prediction.

\paragraph{(2) A harder setting than traditional multi-class RE.}
We argue that this support-based few-shot RE setting is more challenging than
traditional multi-class relation classification for several reasons.

\textbf{Absence of explicit label semantics.} In traditional multi-class RE, relation labels and their semantic descriptions are explicitly provided. In contrast, in our setting the relation semantics are not given directly and must be inferred from the support sentences. As a result, the model must first abstract a relational concept from the support example and then compare it with the concept inferred from the test sentence. This leads to two common failure modes discussed in Section~\ref{sec:error_analysis} (Error Analysis):
(i) failure to focus on relation-specific semantic cues, and
(ii) interdependent errors during relation abstraction across sentences.

\textbf{Local rather than global relation semantics.} In traditional multi-class RE, the model has access to the entire label set, which facilitates fine-grained discrimination between relations. In contrast, our setting exposes the model only to a single support example at a time, resulting in local and incomplete relation semantics.
For example, consider the relations \textit{org:country\_of\_headquarters} and \textit{org:city\_of\_headquarters}. In a traditional multi-class setup, both labels are present, enabling direct comparison. However, in our setting the model may only observe a support sentence expressing \textit{org:country\_of\_headquarters}. When the test sentence expresses \textit{org:city\_of\_headquarters}, the model lacks explicit label definitions or global context, and thus tends to abstract a coarse concept such as \textit{place\_of\_headquarters}, leading to confusion between the two.

\textbf{Empirical evidence of increased difficulty.} This support-based matching formulation has been shown to be challenging in prior work, where overall performance is reported to be low~\cite{alam2024realisticfewshotrelationextraction}. As shown in Appendix~\ref{appendix:inference_results_cross_models} and Table~\ref{tab:inference_results_across_models}, we evaluate a wide range of models under this setting, including 3B--7B models (Qwen2.5-3B/7B, Phi-4-mini-Instruct, Llama-3.1-8B-Instruct, Mistral-7B-Instruct-v0.2) and large mainstream LLMs (DeepSeek-V3, GPT-4o, GPT-3.5-Turbo). We observe that these models perform poorly under our setting when using either \textit{Random Reasoning} or \textit{Direct Matching}. Moreover, reasoning-based generation often increases over-confidence, leading to more false positives. These results demonstrate that realistic few-shot RE remains challenging for current LLMs.

\subsection{Human Evaluation Rubric}\label{appendix:human_evaluation_rubric}

\paragraph{Main Rubric.}To assess the quality of model-generated explanations, we adopt a human evaluation rubric that emphasizes both concise summarization and alignment with RE labeling. 
The rubric assigns scores based on three major criteria: (1) the correctness and conciseness of the summarization for the support sentence, ensuring that the key relational information is accurately captured without including irrelevant details; (2) the correctness and conciseness of the summarization for the test sentence, evaluated under the same principles; and (3) the alignment of the explanation with the labeling of whether the relations expressed in the two sentences match or not, with any abstraction error or illogical reasoning resulting in a deduction.
Each explanation is scored on a 3-point scale, with details provided as follows.

\makebox[\linewidth]{\rule{\linewidth}{0.4pt}}
\textit{Human evaluation rubric for explanation quality (max score: 3).}
\begin{lstlisting}[basicstyle=\small\ttfamily, breaklines=true, breakindent=0em, 
  commentstyle=\color{red!50!green!50!blue!50}, frame=shadowbox, 
  rulesepcolor=\color{red!20!green!20!blue!20}, numbers=none]

[1 point] 
    A correct and concise summarization of the support sentence is awarded 1 point.

[1 point] 
    A correct and concise summarization of the test sentence is awarded 1 point.

[1 point] 
    A reasonable explanation of whether relation_1 and relation_2 match or do not match is awarded 1 point. Any abstraction error results in the loss of this point.

 Special handling:
    - If summarization is incorrect but the explanation is logically
    reasonable, the third point can still be awarded.
    - Points are deducted when the model confuses similar relations,
      e.g., "city" vs. "country", or "reference" vs. "alternate_name".
      
\end{lstlisting}
\makebox[\linewidth]{\rule{\linewidth}{0.4pt}}

\paragraph{Common types of abstraction errors.}In addition to the main rubric, we present several common types of abstraction errors to help graders develop a clearer understanding of how such errors should be identified and penalized. These error types serve as practical references to ensure consistent and fair scoring. The detailed description and illustrative examples are provided for each case as follows.

\makebox[\linewidth]{\rule{\linewidth}{0.4pt}}
\textit{Common Abstraction Errors for Human Rating}
\begin{lstlisting}[basicstyle=\small\ttfamily, breaklines=true, breakindent=0em, 
  commentstyle=\color{red!50!green!50!blue!50}, frame=shadowbox, 
  rulesepcolor=\color{red!20!green!20!blue!20}, numbers=none]

Abstraction Error:
    If two sentences express the same relation, they must be abstracted in the same direction and at the same level.  
    Example:
    ``He, 12-years-old, got a good offer.'' and ``Jam is 12.''  
    - Correct: per:age; per:age  [0 points]  
    - Incorrect: per:age; per:number  [-1 point] 

several common types:
    - Lack of Higher-Level Deductive Abstraction:
        Description: When a higher-level deductive abstraction is required to align the relations, failing to apply it leads to error.  
        Example:
            ``STX Finland is part of the international STX Europe Group'' and  
            ``Merck will acquire all of Millipore.''  
            - Correct: org:parents; org:parents  [0 points]  
            - Incorrect: org:parents; org:transaction  [-1 point]  
        
    - Over- or Under-Focusing on Details:
        Description: The abstraction direction and level are correct, but the model misjudges due to being overly detailed or overly general.
        Example:
            ``The arrangement of financing for Millipore Corp in the US'' and  
            ``Burlington Northern Santa Fe Corp is the biggest bet yet on a US economic recovery.''   
            - Correct: org:country_of_headquarters; org:country_of_headquarters  [0 points]  
            - Incorrect: First = org:financial_transaction, Second = org:economic_event  [-1 point] 
 
\end{lstlisting}
\makebox[\linewidth]{\rule{\linewidth}{0.4pt}}

\paragraph{Rubric Reliability Verification via Cohen’s Kappa.}Furthermore, to verify the reliability of our rubric, we engaged two independent annotators. Both annotators had NLP research backgrounds. They first evaluated the explanations in the setting of Phi-4 on the one-shot TACRED dataset. Each annotator followed the rubric, scoring explanations based on the three criteria and abstraction error types. This independent evaluation enabled us to measure the consistency between annotators. Specifically, for the Phi–TACRED setting, we computed the Cohen’s kappa coefficient to quantify inter-annotator agreement. The resulting kappa value was 0.693, indicating substantial agreement. This shows that our rubric is well-defined and practical. Therefore, we consistently adopted this rubric across all subsequent evaluations, including different stages and experimental settings.

\subsection{Human Evaluation Results}\label{appendix:human_evaluation_results}

To further evaluate explanation quality, we conducted human evaluation across three training stages: (i) vanilla LLMs with the \ours framework, (ii) after RL training with the accuracy reward, and (iii) after RL training with the \reward reward. 

We selected Qwen2.5-14B-Instruct and Phi-4 as base models and evaluated them on two datasets: one-shot TACRED and NYT29.  For each LLM–dataset–stage combination, we sampled 40 explanations, with 10 explanations drawn from each of the four categories: true positives, true negatives, false positives, and false negatives. We consistently adopted the rubric we defined in the Appendix~\ref{appendix:human_evaluation_rubric}. Therefore, for each category, the maximum score is 30 points.

We provide the results of human evaluation in Table~\ref{tab:human-eval-results}. 
The results show that the models trained with the \reward reward consistently outperform both untrained models and accuracy-reward–trained models. For example, in the Phi–TACRED setting, the number of correct explanations in the no\_yes and yes\_no categories increases substantially after applying the \reward reward (from 24 to 29 and from 16 to 26, respectively). Similar improvements are observed in the Qwen–TACRED setting, particularly in the yes\_no category (from 18 to 26).

\begin{table*}[t!]
\centering
\setlength{\tabcolsep}{1pt} 
\renewcommand{\arraystretch}{0.2} 
\begin{tabular}{lccc}
\toprule
\multirow{2}{*}{\textbf{Setting / Category}} & \multicolumn{3}{c}{\textbf{Stages}} \\
\cmidrule(lr){2-4}
& \_untrained & \_rl\_ACC.\_reward & \_rl\_\reward\_reward \\
\midrule
\multicolumn{4}{l}{\textit{\textbf{Phi – TACRED}}} \\
no\_yes  & 24 & 26 & 29 \\
yes\_no  & 16 & 18 & 26 \\
yes\_yes & 27 & 27 & 29 \\
no\_no   & 22 & 22 & 27 \\
\midrule
\multicolumn{4}{l}{\textit{\textbf{Phi – NYT29}}} \\
no\_yes  & 18 & 25 & 24 \\
yes\_no  & 11 & 16 & 27 \\
yes\_yes & 19 & 21 & 22 \\
no\_no   & 21 & 25 & 26 \\
\midrule
\multicolumn{4}{l}{\textit{\textbf{Qwen – TACRED}}} \\
no\_yes  & 22 & 22 & 24 \\
yes\_no  & 18 & 18 & 26 \\
yes\_yes & 26 & 26 & 29 \\
no\_no   & 19 & 21 & 25 \\
\midrule
\multicolumn{4}{l}{\textit{\textbf{Qwen – NYT29}}} \\
no\_yes  & 19 & 22 & 20 \\
yes\_no  &  8 &  5 & 22 \\
yes\_yes & 19 & 14 & 21 \\
no\_no   & 13 & 17 & 28 \\
\bottomrule
\end{tabular}
\vskip 0.1 in
\caption{Human evaluation results across four settings (Phi/Qwen × TACRED/NYT29). 
Each cell reports the score of a category in one combination under human evaluation. 
The categories are defined as follows: 
\textit{no\_yes}: the ground truth is No but the model prediction is Yes; 
\textit{yes\_no}: the ground truth is Yes but the model prediction is No; 
\textit{yes\_yes}: both the ground truth and the model prediction are Yes; 
\textit{no\_no}: both the ground truth and the model prediction is No; }
\label{tab:human-eval-results}
\end{table*}

\clearpage
\newpage
\subsection{Dictionary Robustness Analysis}\label{appendix:dictionary_robustness_analysis}
Building this dictionary is straightforward: it is small (each relation includes only about 10 phrases) and can be created either manually from a few true-positive explanations or automatically using small or large LLMs. We performed a drop-in replacement analysis by using Qwen2.5-14B instead of GPT-4o to extract phrases from explanations on the NYT29 dataset. We evaluate the effect from two aspects:

\textbf{(1) Keyword Overlap.}
We compare the overlap rate between keyword sets extracted by GPT-4o and Qwen2.5-14B.
The two models produce highly consistent keyword sets, indicating that the
extraction process is not sensitive to the choice of LLM.

To measure the consistency between the keyword sets generated by GPT-4o and
Qwen2.5-14B, we use the Overlap Coefficient. Given two keyword sets
$K_{\text{GPT4o}}$ and $K_{\text{Qwen}}$, the overlap is defined as:

\vspace{-30pt}

\begin{equation}
\text{Overlap} =
\frac{|K_{\text{GPT4o}} \cap K_{\text{Qwen}}|}
{\min\left(|K_{\text{GPT4o}}|, |K_{\text{Qwen}}|\right)}.
\end{equation}

\vspace{-30pt}

Table~\ref{tab:rebuild_dictionary_with_opensource_models} shows the phrases for the label
\textit{/location/location/contains} separately by Qwen2.5-14B and GPT4o. We list all the phrases for all the labels separately by Qwen2.5-14B and GPT4o in Appendix~\ref{appendix:keywords_comparison}. In this case shown in the table, we have:
\begin{equation}
\lvert K_{\text{GPT4o}} \cap K_{\text{Qwen}} \rvert = 5,
\\
\min\!\left(\lvert K_{\text{GPT4o}} \rvert, \lvert K_{\text{Qwen}} \rvert\right) = 6,
\end{equation}.

Across all 15 relations in the NYT29 training set, the two models produce highly
consistent keyword sets. In total, we observe:
\begin{equation}
\lvert K_{\text{GPT4o}} \cap K_{\text{Qwen}} \rvert = 80,
\end{equation}

\begin{equation}
\min\!\left(\lvert K_{\text{GPT4o}} \rvert, \lvert K_{\text{Qwen}} \rvert\right) = 87,
\end{equation}
which yields an overall overlap of:
\begin{equation}
\text{Overlap} = \frac{80}{87} = 0.91954.
\end{equation}

\paragraph{(2) Performance of Our Method with a New Dictionary.}
We rebuild the entire dictionary using the phrases extracted by Qwen2.5-14B and
re-run the RL training with Acc+\reward. The resulting F1 score remains very close
to the GPT-4o-based setup, showing that our method is robust under drop-in replacement.

\begin{table}[ht]
\centering
\small
\begin{tabular}{p{0.35\linewidth} p{0.55\linewidth}}
\toprule
\textbf{Qwen2.5-14B} & \textbf{GPT4o} \\
\midrule
\rowcolor{Gainsboro}
\textit{/location/location/contains} &
\textit{/location/location/contains} \\
\midrule
\rowcolor{nyellow}
``city'' &
``city'' \\

\rowcolor{nyellow}
``location'' &
``location'' \\

\rowcolor{nyellow}
``province'' &
``provincial'' \\

\rowcolor{nyellow}
``capital'' &
``capital'' \\

\rowcolor{nyellow}
``located'' &
``located'' \\

 & ``contains'' \\

``part'' & \\
\bottomrule
\end{tabular}
\caption{The phrases for the label\colorbox{Gainsboro}{
\textit{/location/location/contains}} separately by Qwen2.5-14B and GPT4o. The \colorbox{nyellow}{yellow} means the same phrases generated by both open-source and close-source models. We list all the phrases for all the labels separately by Qwen2.5-14B and GPT4o in Appendix~\ref{appendix:keywords_comparison}.}
\label{tab:rebuild_dictionary_with_opensource_models}
\end{table}

\columnbreak

\begin{table}[ht]
\centering
\small
\begin{tabular}{lccc}
\toprule
\textbf{Dictionary Source} & \textbf{P} & \textbf{R} & \textbf{F1} \\
\midrule
GPT-4o      & 63.34 & 38.78 & 48.11 \\
Qwen2.5-14B & 45.35 & 51.28 & 48.13 \\
\bottomrule
\end{tabular}
\vspace{5pt}
\caption{Performance of our method with dictionaries built from different LLMs.
Comparable F1 scores indicate robustness to dictionary replacement.}
\end{table}

\clearpage
\newpage
\subsection{Statistics of the Sampled Training Set}\label{appendix:training_set_distribution}

To ensure fairness and representativeness in one-shot relation extraction evaluation, we construct sampled training sets for both one-shot NYT29 and TACRED that preserve the distributional properties of the original datasets. The algorithm of sampling training data is shown in Algorithm~\ref{alg:training_sampling}.

Tables~\ref{tab:positive_negative_distribution_NYT} and \ref{tab:positive_negative_distribution_TACRED} report the ratio of positives to negatives in the training partition of one-shot NYT29 and TACRED, respectively. In these tables, we compare the ratio of positives to negatives, and the proportion of one of relation labels-\textsc{no\_relation} before and after sampling. For NYT29, it can be seen that the ratio of positive items $(r, r)$, negative items with \textsc{no\_relation}, $(r, \textsc{no\_relation})$, and negative items without \textsc{no\_relation}, $(r, r')$ in the sampled set aligns well with the original dataset. A similar pattern holds for TACRED, where the sampled data maintains the same balance across positive and negative categories.

\begin{table*}[t!]
\centering
\setlength{\tabcolsep}{1pt} 
\renewcommand{\arraystretch}{1.3} 
\begin{adjustbox}{width=1\columnwidth,center}
\begin{tabular}{lcc}
\hline
 & \textbf{Original} & \textbf{Sampled} \\
\hline
Positive items ($r,r$) & 71376 & 2670 \\
Negative items with no\_relation ($r,\text{no\_relation}$) & 571560 & 9660 \\
Negative items without no\_relation ($r,r'$) & 285504 & 7670 \\
\hline
\end{tabular}
\end{adjustbox}

\caption{Ratio of positives and negatives on the original training partition of one-shot NYT29 and our sampled version. 
\textit{Positive items} ($r, r$): pairs where both sentences express the same relation $r$. 
\textit{Negative items with no\_relation} ($r, \text{no\_relation}$): pairs where one sentence expresses a relation $r$ and the other is labeled as no\_relation. 
\textit{Negative items without no\_relation} ($r, r'$): pairs where the two sentences express different relations $r$ and $r'$, neither being no\_relation.}\label{tab:positive_negative_distribution_NYT}
\end{table*}

\begin{table*}[t!]
\centering
\setlength{\tabcolsep}{1pt} 
\renewcommand{\arraystretch}{1.3} 
\begin{tabular}{lcc}
\hline
 & \textbf{Original} & \textbf{Sampled} \\
\hline
Positive items ($r,r$) & 7170 & 2583 \\
Negative items with no\_relation ($r,\text{\textit{no\_relation}}$) & 732075 & 14834 \\
Negative items without no\_relation ($r,r'$) & 28680 & 2583 \\
\hline
\end{tabular}

\caption{Ratio of positives and negatives on the original training partition of one-shot TACRED and our sampled version. 
\textit{Positive items} ($r, r$): pairs where both sentences express the same relation $r$. 
\textit{Negative items with no\_relation} ($r, \text{no\_relation}$): pairs where one sentence expresses a relation $r$ and the other is labeled as no\_relation. 
\textit{Negative items without no\_relation} ($r, r'$): pairs where the two sentences express different relations $r$ and $r'$, neither being no\_relation.}\label{tab:positive_negative_distribution_TACRED}
\end{table*}

\begin{table*}[t!]
\centering
\setlength{\tabcolsep}{1pt} 
\renewcommand{\arraystretch}{1.2} 
\begin{tabular}{l r r}
\hline
\textbf{Relation} & \textbf{Original} & \textbf{Sampled} \\
\hline
/business/company/founders & 52,201 & 1,634 \\
/business/company/place\_founded & 51,408 & 1,571 \\
/business/person/company & 66,753 & 2,226 \\
/film/film\_location/featured\_in\_films & 50,425 & 1,590 \\
/location/country/capital & 70,537 & 2,351 \\
/location/location/contains & 135,244 & 4,934 \\
/location/us\_county/county\_seat & 50,394 & 1,572 \\
/location/us\_state/capital & 51,938 & 1,681 \\
/people/deceased\_person/place\_of\_burial & 49,984 & 1,545 \\
/people/ethnicity/geographic\_distribution & 51,557 & 1,608 \\
/people/person/children & 51,332 & 1,553 \\
/people/person/ethnicity & 50,625 & 1,566 \\
/people/person/nationality & 74,607 & 2,511 \\
/people/person/place\_lived & 71,195 & 2,437 \\
/people/place\_of\_interment/interred\_here & 50,240 & 1,561 \\
no\_relation & 571,560 & 9,660 \\

\hline
\end{tabular}

\caption{Distribution of relation labels in the one-shot NYT29 training partition (original vs. sampled). 
Each row corresponds to a relation type, where the \textit{Original} column reports the number of instances in the full dataset and the \textit{Sampled} column reports the number of instances included in our one-shot sampled version. 
The relation \textsc{no\_relation} indicates sentence pairs that do not express any annotated relation.}\label{tab:label_distribution_NYT}
\end{table*}

\begin{table*}[t!]
\centering
\setlength{\tabcolsep}{1pt} 
\renewcommand{\arraystretch}{1.1} 
\begin{tabular}{l r r}
\hline
\textbf{Relation} & \textbf{Original} & \textbf{Sampled} \\
\hline
org:alternate\_names & 31,532 & 1,156 \\
org:city\_of\_headquarters & 31,016 & 997 \\
org:dissolved & 30,011 & 912 \\
org:members & 30,038 & 968 \\
org:number\_of\_employees/members & 30,236 & 929 \\
org:political/religious\_affiliation & 30,633 & 938 \\
org:shareholders & 30,288 & 945 \\
org:stateorprovince\_of\_headquarters & 30,702 & 1,011 \\
org:subsidiaries & 30,279 & 988 \\
org:website & 30,307 & 946 \\
per:cause\_of\_death & 30,388 & 947 \\
per:charges & 29,975 & 930 \\
per:cities\_of\_residence & 30,833 & 1,003 \\
per:city\_of\_birth & 30,118 & 912 \\
per:countries\_of\_residence & 30,870 & 1,035 \\
per:country\_of\_birth & 29,958 & 907 \\
per:country\_of\_death & 29,833 & 903 \\
per:date\_of\_death & 30,644 & 940 \\
per:employee\_of & 32,941 & 1,383 \\
per:other\_family & 30,106 & 976 \\
per:parents & 30,290 & 959 \\
per:religion & 30,145 & 935 \\
per:spouse & 30,576 & 967 \\
per:stateorprovince\_of\_birth & 30,293 & 908 \\
per:title & 35,913 & 1,671 \\
no\_relation & 732,075 & 14,834 \\
\hline
\end{tabular}

\caption{Distribution of relation labels in the one-shot TACRED training partition (original vs. sampled). 
Each row corresponds to a relation type, where the \textit{Original} column reports the number of instances in the full dataset and the \textit{Sampled} column reports the number of instances included in our one-shot sampled version. 
The relation \textsc{no\_relation} indicates sentence pairs that do not express any annotated relation.}\label{tab:label_distribution_TACRED}
\end{table*}

\begin{algorithm*}[t!]
\caption{Sampling Procedure for Training Data}
\label{alg:sampling}
\begin{algorithmic}[1]
\Require Original dataset $\mathcal{D}$; 
sampling quotas $Q$; 
maximum positives per label $K$
\Ensure Sampled dataset $\mathcal{D}^\prime$
\State Split $\mathcal{D}$ into subsets:
\begin{itemize}
  \item $\mathcal{D}_{r,r}$: $ss\_relation = ts\_relation$ \Comment{Positive pairs}
  \item $\mathcal{D}_{r,no}$: $ts\_relation = \text{no\_relation}$ \Comment{One relation + no\_relation}
  \item $\mathcal{D}_{r,r'}$: $ss\_relation \neq ts\_relation,\ ts\_relation \neq \text{no\_relation}$ \Comment{Different relations}
\end{itemize}
\State $\mathcal{D}^\prime \gets \emptyset$
\State From $\mathcal{D}_{r,r}$: if a relation has more than $K$ pairs, randomly down-sample to $K$; otherwise keep all. Add to $\mathcal{D}^\prime$.
\State From $\mathcal{D}_{r,no}$: group by $ss\_relation$. For each relation $r$, sample $Q[r]$ pairs (or all if fewer available). Add to $\mathcal{D}^\prime$.
\State From $\mathcal{D}_{r,r'}$: randomly sample $2,583$ pairs, approximately preserving label distribution. Add to $\mathcal{D}^\prime$.
\State Shuffle $\mathcal{D}^\prime$.
\State Report statistics: $|\mathcal{D}^\prime|$, \#positives, \#negatives, and ratio.
\State \Return $\mathcal{D}^\prime$
\end{algorithmic}\label{alg:training_sampling}
\end{algorithm*}

In addition, Tables~\ref{tab:label_distribution_NYT} and \ref{tab:label_distribution_TACRED} further analyze the distribution of all the relation labels in the original dataset and the sampled dataset, respectively. Since we sampled the training data strictly according to the original distribution of relation labels, the sampled training datasets have a similar label distribution to the original ones.
These statistics show that our sampled training datasets faithfully reflect the statistical properties of the original datasets, thereby avoiding biases introduced by over- or under-sampling specific relations.

\clearpage
\newpage
\subsection{Statistics of the Sampled Testing Set}\label{appendix:testing_set_distribution}

For the testing partition, we adopt a different sampling strategy from the training data, using a random sampling approach. The resulting sampled testing set preserves the same distributional properties as the original dataset. 
Specifically, (1) the ratio of positive to negative instances remains consistent, (2) the distribution of relation labels is well aligned, and (3) the proportion of \textsc{no\_relation} instances is maintained. These results confirm that our random sampling strategy produces a representative testing partition that faithfully reflects the characteristics of the original testing dataset.

\begin{table*}[t!]
\centering
\setlength{\tabcolsep}{2pt} 
\renewcommand{\arraystretch}{1} 
\begin{tabular}{lcc}
\hline
\textbf{Category} & \textbf{Original} & \textbf{Sampled} \\
\hline
Positive items ($r,r$) & 7,055 & 704 \\
Negative items with no\_relation ($r,\text{\textit{no\_relation}}$) & 114,725 & 11,480 \\
Negative items without no\_relation ($r,r'$) & 28,220 & 2,816 \\
\hline
\end{tabular}
\caption{Ratio of positives and negatives on the original testing partition of one-shot NYT29 and our sampled version. 
\textit{Positive items} ($r, r$): pairs where both sentences express the same relation $r$. 
\textit{Negative items with no\_relation} ($r, \text{no\_relation}$): pairs where one sentence expresses a relation $r$ and the other is labeled as no\_relation. 
\textit{Negative items without no\_relation} ($r, r'$): pairs where the two sentences express different relations $r$ and $r'$, neither being no\_relation.}
\label{tab:nyt_test_pos_neg_distribution}
\end{table*}

\begin{table*}[t!]
\centering
\setlength{\tabcolsep}{1pt} 
\renewcommand{\arraystretch}{1} 
\begin{tabular}{lcc}
\hline
\textbf{Category} & \textbf{Original} & \textbf{Sampled} \\
\hline
Positive items ($r,r$) & 772 & 82 \\
Negative items with no\_relation ($r,\text{\textit{no\_relation}}$) & 146,140 & 14,590 \\
Negative items without no\_relation ($r,r'$) & 3,088 & 328 \\
\hline
\end{tabular}
\caption{Ratio of positives and negatives on the original testing partition of one-shot TACRED and our sampled version. 
\textit{Positive items} ($r, r$): pairs where both sentences express the same relation $r$. 
\textit{Negative items with no\_relation} ($r, \text{no\_relation}$): pairs where one sentence expresses a relation $r$ and the other is labeled as no\_relation. 
\textit{Negative items without no\_relation} ($r, r'$): pairs where the two sentences express different relations $r$ and $r'$, neither being no\_relation.}
\label{tab:tacred_test_pos_neg_distribution}
\end{table*}

\begin{table*}[t!]
\centering
\setlength{\tabcolsep}{1pt} 
\renewcommand{\arraystretch}{1} 
\begin{tabular}{lcc}
\hline
\textbf{Category} & \textbf{Original} & \textbf{Sampled} \\
\hline
Positive items ($r,r$) & 526 & 51 \\
Negative items with no\_relation ($r,\text{\textit{no\_relation}}$) & 147,370 & 14,745 \\
Negative items without no\_relation ($r,r'$) &2,104 & 204 \\
\hline
\end{tabular}
\caption{Ratio of positives and negatives on the original testing partition of one-shot WIKIDATA and our sampled version. 
\textit{Positive items} ($r, r$): pairs where both sentences express the same relation $r$. 
\textit{Negative items with no\_relation} ($r, \text{no\_relation}$): pairs where one sentence expresses a relation $r$ and the other is labeled as no\_relation. 
\textit{Negative items without no\_relation} ($r, r'$): pairs where the two sentences express different relations $r$ and $r'$, neither being no\_relation.}
\label{tab:wikidata_test_pos_neg_distribution}
\end{table*}

\begin{table*}[t!]
\centering
\setlength{\tabcolsep}{1pt} 
\renewcommand{\arraystretch}{1} 
\begin{tabular}{lcc}
\hline
\textbf{Relation} & \textbf{Original} & \textbf{Sampled} \\
\hline
/business/company/major\_shareholders & 30,510 & 3,055 \\
/location/administrative\_division/country & 40,690 & 4,130 \\
/location/country/administrative\_divisions & 41,160 & 4,040 \\
/location/neighborhood/neighborhood\_of & 39,520 & 3,960 \\
/people/deceased\_person/place\_of\_death & 33,395 & 3,335 \\
no\_relation & 114,725 & 11,480 \\
\hline
\end{tabular}
\vskip 0.07 in
\caption{Distribution of relation labels in the one-shot NYT29 testing partition (original vs. sampled). 
Each row corresponds to a relation type, where the \textit{Original} column reports the number of instances in the full dataset and the \textit{Sampled} column reports the number of instances included in our one-shot sampled version. 
The relation \textsc{no\_relation} indicates sentence pairs that do not express any relation.}
\label{tab:nyt29_test_label_distribution}
\end{table*}

\begin{table*}[t!]
\centering
\setlength{\tabcolsep}{1pt} 
\renewcommand{\arraystretch}{1.1} 
\begin{tabular}{lcc}
\hline
\textbf{Relation} & \textbf{Original} & \textbf{Sampled} \\
\hline
no\_relation & 146,140 & 14,590 \\
org:founded\_by & 15,442 & 1,611 \\
org:member\_of & 15,246 & 1,557 \\
org:top\_members/employees & 16,431 & 1,648 \\
per:children & 14,977 & 1,413 \\
per:city\_of\_death & 14,944 & 1,535 \\
per:date\_of\_birth & 15,127 & 1,528 \\
per:origin & 15,513 & 1,582 \\
per:schools\_attended & 15,148 & 1,449 \\
per:siblings & 15,333 & 1,519 \\
per:stateorprovinces\_of\_residence & 15,699 & 1,568 \\
\hline
\end{tabular}
\caption{Distribution of relation labels in the one-shot TACRED testing partition (original vs. sampled).
Each row corresponds to a relation type, where the \textit{Original} column reports the number of instances in the full dataset and the \textit{Sampled} column reports the number of instances included in our one-shot sampled version. 
The relation \textsc{no\_relation} indicates sentence pairs that do not express any annotated relation.}
\label{tab:tacred_test_label_relation_distribution}
\end{table*}

\begin{table*}[t!]
\centering
\setlength{\tabcolsep}{1pt} 
\renewcommand{\arraystretch}{1.1} 
\begin{tabular}{lcc}
\hline
\textbf{Relation} & \textbf{Original} & \textbf{Sampled} \\
\hline
Fach & 1743 & 171 \\
IUCN conservation status & 1653 & 165 \\
Solid solution series with & 1758 & 141 \\
affiliation & 1656 & 165 \\
airline alliance & 1617 & 150 \\
airline hub & 1790 & 174 \\
ancestral home & 1686 & 132 \\
antiparticle & 1629 & 192 \\
architectural style & 1653 & 198 \\
arterial supply & 1731 & 126 \\
based on & 1626 & 138 \\
\hline
\end{tabular}
\caption{ Distribution of relation labels in the one-shot WIKIDATA testing partition (original vs. sampled).
Each row corresponds to a relation type, where the \textit{Original} column reports the number of instances in the full dataset and the \textit{Sampled} column reports the number of instances included in our one-shot sampled version. 
The relation \textsc{no\_relation} indicates sentence pairs that do not express any annotated relation. \textbf{The number of relation labels in WIKIDATA is extremely large, so we split it into three parts. This is the first part.}}
\label{tab:wiki_test_label_relation_distribution_1}
\end{table*}

\clearpage
\newpage

\begin{table}[ht]
\centering
\small
\setlength{\tabcolsep}{1pt} 
\renewcommand{\arraystretch}{1.0}
\begin{tabular}{lcc}
\hline
\textbf{Relation} & \textbf{Original} & \textbf{Sampled} \\
\hline
brother & 1685 & 183 \\
canonization status & 1575 & 162 \\
capital & 1717 & 185 \\
carries & 1698 & 189 \\
cast member & 1802 & 170 \\
composer & 1708 & 156 \\
conferred by & 1688 & 183 \\
connecting line & 1742 & 216 \\
contributor & 1617 & 162 \\
convicted of & 1698 & 150 \\
country & 2481 & 298 \\
country for sport & 1679 & 162 \\
country of origin & 1833 & 204 \\
creator & 1804 & 192 \\
culture & 1584 & 144 \\
depicts & 1644 & 153 \\
describes the fictional universe & 1692 & 132 \\
director & 1761 & 149 \\
director of photography & 1695 & 186 \\
drafted by & 1616 & 183 \\
editor & 1626 & 144 \\
electoral district & 1665 & 123 \\
employer & 1614 & 150 \\
enclave within & 1719 & 120 \\
fabrication method & 1707 & 171 \\
father & 1735 & 135 \\
field of work & 1682 & 159 \\
foundational text & 1755 & 195 \\
founder & 1641 & 177 \\
from fictional universe & 1675 & 156 \\
genre & 1873 & 224 \\
geography of topic & 1731 & 183 \\
given name & 1675 & 150 \\
has cause & 1590 & 165 \\
has quality & 1573 & 141 \\
head of state & 1682 & 168 \\
highest judicial authority & 1755 & 162 \\
influenced by & 1583 & 141 \\
instance of & 2186 & 193 \\
instrumentation & 1743 & 177 \\
is a list of & 1761 & 180 \\
\hline
\end{tabular}
\caption{Distribution of relation labels in the one-shot WIKIDATA testing partition (original vs. sampled), part 2 of 3.}
\label{tab:wiki_test_label_relation_distribution_2}
\end{table}

\begin{table}[ht]
\centering
\small
\setlength{\tabcolsep}{1pt} 
\renewcommand{\arraystretch}{1.0}
\begin{tabular}{lcc}
\hline
\textbf{Relation} & \textbf{Original} & \textbf{Sampled} \\
\hline
item operated & 1776 & 171 \\
killed by & 1821 & 201 \\
license & 1700 & 147 \\
list of episodes & 1626 & 162 \\
located next to body of water & 1752 & 179 \\
lyrics by & 1775 & 213 \\
manifestation of & 1647 & 156 \\
member of political party & 1623 & 177 \\
military branch & 1713 & 159 \\
minor planet group & 1653 & 150 \\
mouth of the watercourse & 1659 & 204 \\
movement & 1767 & 171 \\
no\_relation & 147370 & 14745 \\
occupant & 1697 & 162 \\
office contested & 1721 & 174 \\
official language & 1694 & 171 \\
organizer & 1751 & 174 \\
original network & 1609 & 150 \\
place of birth & 1595 & 189 \\
place of publication & 1731 & 186 \\
points/goal scored by & 1770 & 207 \\
political ideology & 1725 & 210 \\
position played on team / speciality & 1720 & 197 \\
presenter & 1653 & 144 \\
producer & 1755 & 189 \\
product & 1662 & 171 \\
professional or sports partner & 1632 & 171 \\
programming language & 1748 & 174 \\
publication date & 1817 & 204 \\
published in & 1656 & 162 \\
radio format & 1686 & 156 \\
said to be the same as & 1568 & 171 \\
sex or gender & 1731 & 150 \\
student & 1771 & 168 \\
subclass of & 2019 & 220 \\
surface played on & 1734 & 219 \\
track gauge & 1665 & 144 \\
winner & 1776 & 177 \\
\hline
\end{tabular}
\caption{Distribution of relation labels in the one-shot WIKIDATA testing partition (original vs. sampled), part 3 of 3.}
\label{tab:wiki_test_label_relation_distribution_3}
\end{table}

\clearpage
\newpage
\subsection{Prompt for \ours framework}\label{appendix:prompt_CogRE}

\begin{promptbox}[\ours Prompt]{nblue}{co}
You are given two sentences. Follow the three steps below to determine whether they express a similar relation.\\
\\
---\\
\\
Summarization: Focus on the main parts between subjects and objects in the sentences.\\
Summarization examples:\\
\\
Summarize the relations between ``Malcolm Peeler'' and ``Pangburn'' in ``Dr. Malcolm Peeler , grew in Pangburn, has continued the family tradition of practicing medicine in Jonesboro .''.\\
Summarization: Malcolm Peeler came from Pangburn.\\

Summarize the relations between ``Oceania'' and ``PECC'' in ``Oceania and the Western Hemisphere within the PECC region , as surplus food producers and exporters , confront unique consumer issues , such as lower food expenditure and higher caloric intake compared to Asia .''.\\
Summarization: Oceania within region PECC.\\

Summarize the relations between ``Global Climate Research Institute'' and ``GCRI'' in ``Climate change challenges remain a key concern at the annual summit. The outlook is concerning, according to the Global Climate Research Institute ( GCRI ), which coordinates the event each year.''.\\
Summarization: Global Climate Research Institute is abbreviated as GCRI.\\

Summarize the relations between ``Panasonic Corp'' and ``Tesla Inc'' in ``Tesla Inc. is a wholly-owned subsidiary of Panasonic Corp, focusing on energy storage solutions.''.\\
Summarization: Panasonic Corp is a subsidiary of Tesla Inc.\\
\\
---\\
\\
Step 1: summarize the relations between ``\{support\_sentence\_subject\}'' and \\ 
``\{support\_sentence\_object\}'' in ``\{support\_sentence\}''.\\
Label your result as: Relation\_Summarization\_1.\\

Step 2: summarize the relations between ``\{test\_sentence\_subject\}'' and \\
``\{test\_sentence\_object\}'' in ``\{test\_sentence\}''.\\
Label your result as: Relation\_Summarization\_2.\\

Step 3: are the relations between ``\{support\_sentence\_subject\}'' and \\
``\{support\_sentence\_object\}'' in Relation\_Summarization\_1 and between ``\{test\_sentence\_subject\}'' and ``\{test\_sentence\_object\}'' in Relation\_Summarization\_2 similar? \\
Focus on the phrases in the Relation\_Summarization\_1 an Relation\_Summarization\_2 that convey relations.\\
\\
Generate the understanding process, followed by Yes or No in a separate line.\\

\end{promptbox}

\subsection{Prompt for Relation Phrase Extraction}\label{appendix:Prompt_for_Relation_Keyword_Extraction}

\begin{promptbox}[Phrases Extraction Prompt for GPT-4o]{nblue}{ek}
Relation: \{relation\}\\
Please extract the words or phrases that indicate trigger words or relation summaries from the following answers; the relation is \{relation\}.\\
Output a string list contain all the words.\\
output\_case\_1:\\
\{content\_1\}\\
support\_sentence: \{support\_sentence\_1\}\\
test\_sentence: \{test\_sentence\_1\}\\
\\
output\_case\_2:\\
\{content\_2\}\\
support\_sentence: \{support\_sentence\_2\}\\
test\_sentence: \{test\_sentence\_2\}\\
\\
output\_case\_3:\\
\{content\_3\}\\
support\_sentence: \{support\_sentence\_3\}\\
test\_sentence: \{test\_sentence\_3\}\\
\\
output\_case\_4:\\
\{content\_4\}\\
support\_sentence: \{support\_sentence\_4\}\\
test\_sentence: \{test\_sentence\_4\}\\
\\
output\_case\_5:\\
\{content\_5\}\\
support\_sentence: \{support\_sentence\_5\}\\
test\_sentence: \{test\_sentence\_5\}\\
\\
\end{promptbox}

\subsection{Prompt for Baselines}\label{appendix:naive_prompt}

\subsubsection{Prompts for SUMASK}\label{appendix:SUMASK_prompt}

\begin{promptbox}[Realistic FSRE SUMASK Prompt]{nblue}{sa}
You are given two sentences. Follow the three steps below to determine whether they express a similar relation. \\
\\ 
--- \\ 
\\
Summarization examples:\\
\\
Summarize the relations between ``Malcolm Peeler'' and ``Pangburn'' in ``Dr. Malcolm Peeler , grew in Pangburn, has continued the family tradition of practicing medicine in Jonesboro .''.\\
Summarization: Malcolm Peeler came from Pangburn.\\
\\
Summarize the relations between ``Oceania'' and ``PECC'' in ``Oceania and the Western Hemisphere within the PECC region , as surplus food producers and exporters , confront unique consumer issues , such as lower food expenditure and higher caloric intake compared to Asia .''.\\
Summarization: Oceania within region PECC.\\
\\
Summarize the relations between ``Global Climate Research Institute'' and ``GCRI'' in ``Climate change challenges remain a key concern at the annual summit. The outlook is concerning, according to the Global Climate Research Institute ( GCRI ), which coordinates the event each year.''.\\
Summarization: Global Climate Research Institute is abbreviated as GCRI.\\

Summarize the relations between ``Panasonic Corp'' and ``Tesla Inc'' in ``Tesla Inc. is a wholly-owned subsidiary of Panasonic Corp, focusing on energy storage solutions.''.\\
Summarization: Panasonic Corp is a subsidiary of Tesla Inc.\\
\\
---\\
\\
Step 1: summarize the relations between ``\{support\_sentence\_subject\}'' and \\
``\{support\_sentence\_object\}'' in ``\{support\_sentence\}''.\\
Label your result as: Relation\_Summarization\_1.\\
\\
Step 2: summarize the relations between ``\{test\_sentence\_subject\}'' and \\
``\{test\_sentence\_object\}'' in ``\{test\_sentence\}''.\\
Label your result as: Relation\_Summarization\_2.\\
\\
Step 3: generate a question as: are the relations between ``\{support\_sentence\_subject\}'' and ``\{support\_sentence\_object\}'' in Relation\_Summarization\_1 and between ``\{test\_sentence\_subject\}'' and ``\{test\_sentence\_object\}'' in Relation\_Summarization\_2 similar? \\
\\
Step 4: directly answer the question with Yes or No in a separate line.
\end{promptbox}
\subsubsection{Direct-Answer Prompt}\label{appendix:Direct-Answer-Prompt}

\begin{promptbox}[Direct-Answer Prompt]{nblue}{da}
Are the relations between ``\{support\_sentence\_subject\}'' and ``\{support\_sentence\_object\}'' in ``\{support\_sentence\}'' and between ``\{test\_sentence\_subject\}'' and ``\{test\_sentence\_object\}'' in \{test\_sentence\} similar? 
Directly answer Yes or No in a separate line.\\
---\\
IMPORTANT: must answer with just Yes or No.
\end{promptbox}
\subsubsection{Simple-Reasoning Prompt}\label{appendix:Simple-Reasoning-Prompt}

\begin{promptbox}[Simple-Reasoning Prompt]{nblue}{sr}
Are the relations between ``\{support\_sentence\_subject\}'' and ``\{support\_sentence\_object\}'' in ``\{support\_sentence\}'' and between ``\{test\_sentence\_subject\}'' and ``\{test\_sentence\_object\}'' in \{test\_sentence\} similar? \\
Generate the understanding process, followed by Yes or No in a separate line.

\end{promptbox}

\subsection{Prompt for Ablation experiments}\label{appendix:prompt_for_ablation_experiments}
\subsubsection{COGRE- w/o chunking Prompt}\label{appendix:prompt_wo_chunking}

\begin{promptbox}[\ours- \textit{w/o} chunking Prompt]{nblue}{wc}
You are given two sentences. Follow the three steps below to determine whether they express a similar relation.\\
Step1: are the relations between ``\{paraphrased\_sentence\_subject\}'' and \\
``\{paraphrased\_sentence\_object\}'' in \{paraphrased\_sentence\} and between ``\{test\_sentence\_subject\}'' and ``\{test\_sentence\_object\}'' in \{test\_sentence\} similar? Focus on the phrases in the \{paraphrased\_sentence\} an \{test\_sentence\} that convey relations.\\
Generate the understanding process, followed by Yes or No in a separate line.\\
\\
\end{promptbox}
\subsubsection{COGRE- w/o reasoning Prompt}\label{appendix:prompt_wo_reasoning}

\begin{promptbox}[\ours- \textit{w/o} reasoning Prompt]{nblue}{wr}
You are given two sentences. Follow the three steps below to determine whether they express a similar relation.\\
---\\
Summarization examples:\\
\\
Summarize the relations between ``Malcolm Peeler'' and ``Pangburn'' in ``Dr. Malcolm Peeler , grew in Pangburn, has continued the family tradition of practicing medicine in Jonesboro .''.\\
Summarization: Malcolm Peeler came from Pangburn.\\
\\
Summarize the relations between ``Oceania'' and ``PECC'' in ``Oceania and the Western Hemisphere within the PECC region , as surplus food producers and exporters , confront unique consumer issues , such as lower food expenditure and higher caloric intake compared to Asia .''.\\
Summarization: Oceania within region PECC.\\
\\
Summarize the relations between ``Global Climate Research Institute'' and ``GCRI'' in ``Climate change challenges remain a key concern at the annual summit. The outlook is concerning, according to the Global Climate Research Institute ( GCRI ), which coordinates the event each year.''.\\
Summarization: Global Climate Research Institute is abbreviated as GCRI.\\
\\
Summarize the relations between ``Panasonic Corp'' and ``Tesla Inc'' in ``Tesla Inc. is a wholly-owned subsidiary of Panasonic Corp, focusing on energy storage solutions.''.\\
Summarization: Panasonic Corp is a subsidiary of Tesla Inc.\\
---\\
Step 1: summarize the relations between ``\{support\_sentence\_subject\}'' and \\
``\{support\_sentence\_object\}'' in ``\{support\_sentence\}''.\\
Label your result as: Relation\_Summarization\_1.\\
\\
Step 2: summarize the relations between ``\{test\_sentence\_subject\}'' and ``\{test\_sentence\_object\}'' in ``\{test\_sentence\}''.\\
Label your result as: Relation\_Summarization\_2.\\
\\
Step 3: generate a question as: are the relations between ``\{support\_sentence\_subject\}'' and ``\{support\_sentence\_object\}'' in Relation\_Summarization\_1 and between ``\{test\_sentence\_subject\}'' and ``\{test\_sentence\_object\}'' in Relation\_Summarization\_2 similar? \\
\\
Step 4: Focus on the phrases in the Relation\_Summarization\_1 an Relation\_Summarization\_2 that convey relations, and directly answer the question with Yes or No in a separate line.\\
\\
\end{promptbox}
\subsubsection{COGRE- w/o phrases Prompt}\label{appendix:prompt_wo_keywords}

\begin{promptbox}[\ours- \textit{w/o} phrase Prompt]{nblue}{wk}
You are given two sentences. Follow the three steps below to determine whether they express a similar relation.\\
---\\
Summarization examples:\\
\\
Summarize the relations between ``Malcolm Peeler'' and ``Pangburn'' in ``Dr. Malcolm Peeler , grew in Pangburn, has continued the family tradition of practicing medicine in Jonesboro .''.\\
Summarization: Malcolm Peeler came from Pangburn.\\
\\
Summarize the relations between ``Oceania'' and ``PECC'' in ``Oceania and the Western Hemisphere within the PECC region , as surplus food producers and exporters , confront unique consumer issues , such as lower food expenditure and higher caloric intake compared to Asia .''.\\
Summarization: Oceania within region PECC.\\
\\
Summarize the relations between ``Global Climate Research Institute'' and ``GCRI'' in ``Climate change challenges remain a key concern at the annual summit. The outlook is concerning, according to the Global Climate Research Institute ( GCRI ), which coordinates the event each year.''.
Summarization: Global Climate Research Institute is abbreviated as GCRI.\\
\\
Summarize the relations between ``Panasonic Corp'' and ``Tesla Inc'' in ``Tesla Inc. is a wholly-owned subsidiary of Panasonic Corp, focusing on energy storage solutions.''.\\
Summarization: Panasonic Corp is a subsidiary of Tesla Inc.\\
---\\
Step 1: summarize the relations between ``\{support\_sentence\_subject\}'' and \\
``\{support\_sentence\_object\}'' in ``\{support\_sentence\}''.\\
Label your result as: Relation\_Summarization\_1.\\
\\
Step 2: summarize the relations between ``\{test\_sentence\_subject\}'' and ``\{test\_sentence\_object\}'' in ``\{test\_sentence\}''.\\
Label your result as: Relation\_Summarization\_2.\\
\\
Step 3: generate a question as: are the relations between ``\{support\_sentence\_subject\}'' and ``\{support\_sentence\_object\}'' in Relation\_Summarization\_1 and between ``\{test\_sentence\_subject\}'' and ``\{test\_sentence\_object\}'' in Relation\_Summarization\_2 similar? \\
Generate the understanding process, followed by Yes or No in a separate line.\\
\\
\end{promptbox}

\clearpage

\newpage
\clearpage
\subsection{Inference Results across Model Families and Sizes on realistic One-shot RE task}\label{appendix:inference_results_cross_models}
In addition to Phi-4 and Qwen2.5-14B-Instruct, we further evaluate our \ours reasoning method on models from four additional families of varying sizes. The results, shown in Table~\ref{tab:inference_results_across_models}, reveal two key findings: (1) \ours consistently outperforms prompting-based baselines across both model families and model sizes; and (2) models with fewer than 10B parameters perform poorly on the one-shot RE task.

We also include three mainstream LLMs in Table~\ref{tab:inference_results_across_models} to evaluate the mainstream LLMs' performance in this realistic Few-shot Relation Extraction task.

\begin{table*}[t!]
\centering
\setlength{\tabcolsep}{9pt} 
\renewcommand{\arraystretch}{1} 
\begin{tabular}{p{6cm}ccccccc}
\toprule
\multirow{1}{*}{Technique} 
& \multicolumn{3}{c}{One-shot TACRED} \\
\cmidrule(lr){2-4} 
& P & R & F1 & \\
\midrule
\textit{\textbf{Qwen2.5-7B-Instruct}} \\
Direct Matching             & 22.58 & 9.21 & 13.08   \\
Random Reasoning            & 3.42 & 43.42 & 6.35    \\
Cognitive-Structured RE (\textit{our})       
                            & 10.24 & 50.00 & \colorbox{nblue!50}{17.00}  \\
\midrule
\textit{\textbf{Qwen2.5-3B-Instruct}} \\
Direct Matching             & 100.00 & 0.00 & 0.00   \\
Random Reasoning            & 10.53 & 7.89 & 9.02    \\
Cognitive-Structured RE (\textit{our})       
                            & 4.69 & 36.84 & 8.32  \\
\midrule
\textit{\textbf{Phi-4-mini-Instruct (3.8B)}} \\
Direct Matching             & 0.88 & 28.95 & 1.70   \\
Random Reasoning            & 1.65 & 52.63 & 3.21    \\
Cognitive-Structured RE (\textit{our})       
                            & 9.35 & 34.21 & \colorbox{nblue!50}{14.69}  \\
\midrule
\textit{\textbf{Llama-3.1-8B-Instruct}} \\
Direct Matching             & 1.61 & 39.47 & 3.09   \\
Random Reasoning            & 0.43 & 17.11 & 0.85    \\
Cognitive-Structured RE (\textit{our})       
                            & 0.72 & 27.63 & 1.40  \\
\midrule
\textit{\textbf{Mistral-7B-Instruct-v0.2}} \\
Direct Matching             & 5.00 & 17.11 & 7.74   \\
Random Reasoning            & 0.95 & 26.32 & 1.83    \\
Cognitive-Structured RE (\textit{our})       
                            & 5.14 & 25.00 & \colorbox{nblue!50}{8.52}  \\
\midrule
\textit{\textbf{Deepseek-v3}} \\
Direct Matching             & 35.62 & 37.14 & 36.36   \\
Random Reasoning            & 38.46 & 21.43 & 27.52    \\
\midrule
\textit{\textbf{GPT-4o}} \\
Random Reasoning            & 20.00 & 23.00 & 21.33    \\
\midrule
\textit{\textbf{GPT-3.5-turbo}} \\
Random Reasoning            & 16.23 & 36.00 & 22.32    \\
\bottomrule
\end{tabular}
\vspace{5pt}
\caption{Performance comparison of different model families and sizes on the one-shot TACRED task. 
We evaluate our \ours framework against prompting-based baselines (Direct Matching and Random Reasoning). 
}
\label{tab:inference_results_across_models}
\end{table*}

\newpage
\subsection{Cases Comparison of Phi-4 on TACRED}\label{appendix:cases_comparison_phi}

As discussed in Section~\ref{sec:error_analysis}, we examine the quality of explanations generated by LLMs across three stages: (1) vanilla LLM, (2) LLM trained with RL using only the accuracy reward, and (3) LLM trained with RL using both the accuracy reward and the \reward reward. We conducted evaluations on four model–dataset combinations, sampling and analyzing 40 instances for each. Here, we present examples from Phi-4 on the one-shot TACRED dataset. Specifically, we provide six illustrative cases: in five of them, the LLM trained with both accuracy and \reward rewards produces more concise summaries, and in all six cases, its explanations include relational phrases that align more closely with the gold relation labels. In the following examples, we highlight poor behaviors in yellow and the corresponding improvements in blue.

\begin{table*}[t!]
\centering
\begin{adjustbox}{width=2\columnwidth,center}
\begin{tabular}{p{14cm} }
\hline

\rowcolor{Gainsboro} \multicolumn{1}{c}{\textit{Cases1 - Error Analysis from three Stages LLMs}} \\
$\blacktriangleright$ \textbf{ Input:}\\
1. Federal regulators filed a civil fraud lawsuit Thursday against \textbf{Countrywide Financial} Corp co-founder \textbf{Angelo Mozilo} and two former associates.\\
2. It traveled to Rice University, where de Menil and his wife, \textbf{Dominique de Menil}, who later founded the \textbf{Menil Collection}, ran the art museum.\\
\\

$\blacktriangleright$ \textbf{ Gold Relation Labels and Ground Truth:}\\
1. org:founded\_by \\
2. org:founded\_by \\
3. Yes \\
\\
$\blacktriangleright$\textbf{ Vanilla LLMs Output:} \\
Relation\_Summarization\_1: \\
\colorbox{nyellow!90}{Federal regulators filed a civil fraud lawsuit against Angelo Mozilo, }\\
\colorbox{nyellow!90}{co-founder of Countrywide Financial Corp.}\\
Relation\_Summarization\_2:  
Dominique de Menil founded the Menil Collection.

Understanding Process:  
- In Relation\_Summarization\_1, the relation is between a legal action (lawsuit) and Angelo Mozilo, who is associated with Countrywide Financial Corp as a co-founder.  
- In Relation\_Summarization\_2, the relation is between Dominique de Menil and the Menil Collection, where Dominique is the founder.  
The relations differ in nature: one involves \colorbox{nyellow!90}{a legal action against} a co-founder, and the other involves \colorbox{nyellow!90}{founding an institution}.\\
\\
$\blacktriangleright$ \textbf{ Trained LLMs with Acc. Output:} \\
Relation\_Summarization\_1:  \\
\colorbox{nblue!35}{Angelo Mozilo is a co-founder of Countrywide Financial Corp.}\\

Relation\_Summarization\_2:  \\
Dominique de Menil founded the Menil Collection.\\

Understanding Process:  \\
- In Relation\_Summarization\_1, the relation is between a person (Angelo Mozilo) and a corporation (Countrywide Financial Corp), where the person is identified as a co-founder of the corporation.  \\
- In Relation\_Summarization\_2, the relation is between a person (Dominique de Menil) and a collection (Menil Collection), where the person is identified as the founder of the collection.  \\
- Both summaries describe a person's role in an entity, but the nature of the entities differs: \colorbox{nyellow!90}{one is a corporation}, and the other is\colorbox{nyellow!90}{ a collection}.\\
\\
$\blacktriangleright$ \textbf{ Trained LLMs with \reward Output:} \\
Relation\_Summarization\_1: \\
\colorbox{nblue!35}{Angelo Mozilo is a co-founder of Countrywide Financial Corp.}\\

Relation\_Summarization\_2: Dominique de Menil is the founder of Menil Collection.\\

Understanding Process: \\
In Relation\_Summarization\_1, the relation is organizational, indicating that Angelo Mozilo is a co-founder of Countrywide Financial Corp. \\
In Relation\_Summarization\_2, the relation is organizational, indicating that Dominique de Menil is the founder of Menil Collection. \\
In both cases, the relation is the same: a \colorbox{nblue!35}{person} is the \colorbox{nblue!35}{founder} of a \colorbox{nblue!35}{company}.\\

\bottomrule
\end{tabular}
\end{adjustbox}

\end{table*}

\begin{table*}[htbp]
\centering
\begin{adjustbox}{width=1.6\columnwidth,center}
\begin{tabular}{p{14cm} }
\hline

\rowcolor{Gainsboro} \multicolumn{1}{c}{\textit{Cases2 - Error Analysis from three Stages LLMs}} \\

$\blacktriangleright$\textbf{ Input:}\\
1. The Honduran interim government would not be recognized by the international community, \textbf{Jacinto Suarez}, Nicaraguan deputy to the \textbf{Central American Parliament} ( PARLACEN ) said Monday.\\
2. SEC's McKown to Step Down as Chief Counsel for Enforcement Her replacement probably will inherit pending high-profile cases, including a fraud lawsuit against Goldman Sachs and an insider-trading case against \textbf{Angelo Mozilo}, the former chief executive officer of \textbf{Countrywide Financial} Corp.\\
\\

$\blacktriangleright$\textbf{ Gold Relation Labels and Ground Truth:}\\
1. org:top\_members/employees \\
2. org:top\_members/employees \\
3. Yes \\
\\

$\blacktriangleright$ \textbf{ Vanilla LLMs Output:} \\
Relation\_Summarization\_1:  Jacinto Suarez is a \colorbox{nyellow!90}{Nicaraguan deputy} to the Central American Parliament.\\
Relation\_Summarization\_2:  Angelo Mozilo is the former chief executive officer of Countrywide Financial Corp.\\
Understanding Process:  \\
Relation\_Summarization\_1 describes Jacinto Suarez's role as a deputy to the Central American Parliament, indicating a membership or representative relationship.\\  
Relation\_Summarization\_2 describes Angelo Mozilo's role as the former CEO of Countrywide Financial Corp, indicating a leadership or executive position within the company.  \\
The nature of the relationships differs: one is a \colorbox{nyellow!90}{representative role within a parliamentary body}, and the other is an \colorbox{nyellow!90}{executive role within a corporation}. \\ 
Conclusion: No.\\
\\

$\blacktriangleright$ \textbf{ Trained LLMs with Acc. Output:} \\
Relation\_Summarization\_1:  Jacinto Suarez is a \colorbox{nyellow!90}{Nicaraguan deputy} to the Central American Parliament. \\
Relation\_Summarization\_2:  Angelo Mozilo is the former chief executive officer of Countrywide Financial Corp.\\
Understanding Process:  \\
Relation\_Summarization\_1 identifies Jacinto Suarez as holding a position (deputy) within the Central American Parliament.  \\
Relation\_Summarization\_2 identifies Angelo Mozilo as having held a position (chief executive officer) within Countrywide Financial Corp.  \\
Both summaries describe \colorbox{nyellow!90}{a person and their role or position within an organization.}  \\
The nature of the roles is similar in that both involve being part of an organization, but the specific roles differ (deputy vs. CEO).\\  

Conclusion: No\\
\\

$\blacktriangleright$\textbf{ Trained LLMs with \reward Output:} \\
Relation\_Summarization\_1:  Jacinto Suarez is a \colorbox{nblue!35}{deputy} to the Central American Parliament.\\
Relation\_Summarization\_2:  Angelo Mozilo is the former chief executive officer of Countrywide Financial Corp.\\
Understanding Process:  \\
In Relation\_Summarization\_1, the relation is organizational, indicating that Jacinto Suarez holds the position of deputy to the Central American Parliament.  \\
In Relation\_Summarization\_2, the relation is occupational, indicating that Angelo Mozilo is the former CEO of Countrywide Financial Corp.  \\
In both cases, the relation is the same: a \colorbox{nblue!35}{person} is \colorbox{nblue!35}{employed} in a specific role within a \colorbox{nblue!35}{company}/\colorbox{nblue!35}{organization}.  \\

Yes\\

\bottomrule
\end{tabular}
\end{adjustbox}

\end{table*}

\begin{table*}[htbp]
\centering
\begin{adjustbox}{width=1.8\columnwidth,center}
\begin{tabular}{p{14cm} }
\hline

\rowcolor{Gainsboro} \multicolumn{1}{c}{\textit{Cases3 - Error Analysis from three Stages LLMs}} \\
$\blacktriangleright$\textbf{ Input:}\\
1. The religious re-awakening culminated in the 1998 visit by Catholic Pope John Paul II, but \textbf{Dworin} recalls with pride that the \textbf{Cuban} leader did not ignore the Jews.\\
2. In Port-au-Prince, \textbf{Laura Silsby}, the leader of the 10 \textbf{American} missionaries arrested at the Dominican border for trying to smuggle out a busload of children, appeared in court Monday to answer kidnapping charges.\\
\\

$\blacktriangleright$\textbf{ Gold Relation Labels and Ground Truth:}\\
1. per:origin \\
2. per:origin \\
3. Yes \\
\\
$\blacktriangleright$\textbf{ Vanilla LLMs Output:} \\
Relation\_Summarization\_1:  \colorbox{nyellow!90}{Dworin recalls the Cuban leader acknowledging the Jews.}\\
Relation\_Summarization\_2:  \colorbox{nyellow!90}{Laura Silsby is the leader of the American missionaries.}\\
Understanding Process:  \\
1. Relation\_Summarization\_1 focuses on Dworin's recollection of the Cuban leader's acknowledgment of Jews, indicating a historical or \colorbox{nyellow!90}{cultural recognition}.  \\
2. Relation\_Summarization\_2 describes Laura Silsby's role as the leader of a group of American missionaries, indicating a leadership or organizational relationship.\\ 
The relations are not similar: the first is about historical acknowledgment, while the second is about \colorbox{nyellow!90}{leadership} within a group.\\
\\
$\blacktriangleright$\textbf{ Trained LLMs with Acc. Output:} \\
Relation\_Summarization\_1:
\colorbox{nyellow!90}{Dworin recalls the Cuban leader acknowledging Jews.}\\

Relation\_Summarization\_2: 
\colorbox{nyellow!90}{Laura Silsby is the leader of American missionaries.}\\

Understanding Process:  \\
Relation\_Summarization\_1 highlights a \colorbox{nyellow!90}{historical event} where Dworin recalls a Cuban leader's acknowledgment of Jews, indicating a \colorbox{nyellow!90}{recognition} or interaction.  \\
Relation\_Summarization\_2 describes Laura Silsby's role as the leader of a group of American missionaries, indicating a \colorbox{nyellow!90}{leadership} or organizational relationship.  \\

The phrases in Relation\_Summarization\_1 ("recalls," "acknowledging") suggest a memory or recognition of an event involving a Cuban leader.  \\
The phrases in Relation\_Summarization\_2 ("leader," "American missionaries") suggest a leadership role within a group.  \\

The nature of the relationships is different: one is about \colorbox{nyellow!90}{historical acknowledgment}, and the other is about \colorbox{nyellow!90}{leadership}.\\
\\
$\blacktriangleright$\textbf{ Trained LLMs with \reward Output:} \\
Relation\_Summarization\_1:  \colorbox{nblue!35}{Dworin is Cuban.}\\
Relation\_Summarization\_2:  \colorbox{nblue!35}{Laura Silsby is American.}\\
Understanding Process:  \\
In Relation\_Summarization\_1, the relation is \colorbox{nblue!35}{national identity}, indicating that Dworin is Cuban.  \\
In Relation\_Summarization\_2, the relation is \colorbox{nblue!35}{identity}, indicating that Laura Silsby is American.  \\
In both cases, the relation is the same: A person is \colorbox{nblue!35}{originally} from a country.\\

\bottomrule
\end{tabular}
\end{adjustbox}

\end{table*}

\begin{table*}[htbp]
\centering
\begin{adjustbox}{width=1.8\columnwidth,center}
\begin{tabular}{p{14cm} }
\hline

\rowcolor{Gainsboro} \multicolumn{1}{c}{\textit{Cases4 - Error Analysis from three Stages LLMs}} \\

$\blacktriangleright$\textbf{ Input:}\\
1. This news comes from \textbf{Karr Ingham}, an economist who created the \textbf{Texas Petro Index} ( TPI ), which is a service of the Texas Alliance of Energy Producers.\\
2. \textbf{Ahmed Rashid}, a Pakistani journalist, is the author of ``Descent into Chaos: The United States and the \textbf{Failure of Nation Building} in Afghanistan, Pakistan and Central Asia.''\\
\\

$\blacktriangleright$\textbf{ Gold Relation Labels and Ground Truth:}\\
1. org:founded\_by \\
2. no\_relation \\
3. No \\
\\

$\blacktriangleright$\textbf{ Vanilla LLMs Output:} \\
Relation\_Summarization\_1:  Karr Ingham created the Texas Petro Index.\\
Relation\_Summarization\_2:  Ahmed Rashid is the author of "Failure of Nation Building."\\
Understanding Process:  \\
Relation\_Summarization\_1 identifies Karr Ingham as the creator of the Texas Petro Index.  \\
Relation\_Summarization\_2 identifies Ahmed Rashid as the author of "Failure of Nation Building."  \\
The key relation in both summaries is one of creation or authorship.\\ 
In both cases, the first \colorbox{nyellow!90}{subject} (Karr Ingham and Ahmed Rashid) is responsible for \colorbox{nyellow!90}{creating or authoring} the second \colorbox{nyellow!90}{subject} (Texas Petro Index and "Failure of Nation Building").  \\
Conclusion: Yes.\\
\\

$\blacktriangleright$\textbf{ Trained LLMs with Acc. Output:} \\
Relation\_Summarization\_1:  Karr Ingham created the Texas Petro Index.\\

Relation\_Summarization\_2:  Ahmed Rashid is the author of "Failure of Nation Building."\\

Understanding Process:  \\
Relation\_Summarization\_1 highlights a \colorbox{nyellow!90}{creator-creation} relationship where Karr Ingham is the creator of the Texas Petro Index.  \\
Relation\_Summarization\_2 highlights an \colorbox{nyellow!90}{author-work} relationship where Ahmed Rashid is the author of "Failure of Nation Building."  \\
Both summaries describe a relationship where one \colorbox{nyellow!90}{entity} (Karr Ingham/Ahmed Rashid) is responsible for \colorbox{nyellow!90}{producing or creating} another \colorbox{nyellow!90}{entity} (Texas Petro Index/"Failure of Nation Building").  \\

Conclusion: Yes.\\
\\

$\blacktriangleright$\textbf{ Trained LLMs with \reward Output:} \\
Relation\_Summarization\_1:  Karr Ingham is the creator of the Texas Petro Index.\\
Relation\_Summarization\_2:  Ahmed Rashid is the author of "Failure of Nation Building."\\

Understanding Process:  \\
In Relation\_Summarization\_1, the relation is \colorbox{nblue!35}{organizational}, indicating that Karr Ingham is the person who created the Texas Petro Index.  \\
In Relation\_Summarization\_2, the relation is \colorbox{nblue!35}{authorship}, indicating that Ahmed Rashid is the author of a specific book.  \\
The nature of these relations is different: one is about creation of an index, and the other is about authorship of a book.  \\
No\\

\bottomrule
\end{tabular}
\end{adjustbox}

\end{table*}

\begin{table*}[htbp]
\centering
\begin{adjustbox}{width=1.8\columnwidth,center}
\begin{tabular}{p{14cm} }
\hline

\rowcolor{Gainsboro} \multicolumn{1}{c}{\textit{Cases5 - Error Analysis from three Stages LLMs}} \\

$\blacktriangleright$\textbf{ Input:}\\
1. ( 1:32 ) R. `` William Kunstler : Disturbing the Universe '' -- Filmmakers \textbf{Emily} and \textbf{Sarah Kunstler} explore the life of their father, the late radical civil rights lawyer who fought with Martin Luther King, Jr. and represented the famed ``Chicago 8'' activists who protested the Vietnam War.\\
2. \textbf{He} is also survived by his parents and a sister, Karen Lange, of Washington, and a brother, \textbf{Adam Lange}, of St Louis.\\
\\

$\blacktriangleright$\textbf{ Gold Relation Labels and Ground Truth:}\\
1. per:siblings \\
2. per:siblings \\
3. Yes \\

$\blacktriangleright$\textbf{ Vanilla LLMs Output:} \\
Relation\_Summarization\_1: \\
\colorbox{nyellow!90}{Sarah Kunstler and Emily are filmmakers exploring the life of their father.}\\
Relation\_Summarization\_2:  He is survived by his brother, Adam Lange.\\
Understanding Process:  \\
- In Relation\_Summarization\_1, the relation is between Sarah Kunstler and Emily, who are \colorbox{nyellow!90}{collaborators} in a filmmaking project.\\ 
- In Relation\_Summarization\_2, the relation is between "He" and Adam Lange, who are siblings.  \\
The phrases in Relation\_Summarization\_1 ("filmmakers," "exploring") indicate a \colorbox{nyellow!90}{professional and collaborative} relationship. \\
In Relation\_Summarization\_2 ("survived by," "brother"), the relationship is familial.  \\
Conclusion: No\\

$\blacktriangleright$\textbf{ Trained LLMs with Acc. Output:} \\
Relation\_Summarization\_1: \\
\colorbox{nyellow!90}{Sarah Kunstler and Emily are filmmakers exploring the life of their father.}\\
Relation\_Summarization\_2:  He and Adam Lange are siblings.\\
Understanding Process:  \\
Relation\_Summarization\_1 identifies Sarah Kunstler and Emily as collaborators in a filmmaking project about their father.\\  
Relation\_Summarization\_2 identifies a familial relationship between "He" and Adam Lange as siblings.  \\
The phrases in Relation\_Summarization\_1 are "filmmakers" and "exploring," indicating a \colorbox{nyellow!90}{professional} and \colorbox{nyellow!90}{collaborative} relationship.  \\
The phrases in Relation\_Summarization\_2 are "survived by" and "brother," indicating a familial relationship.  \\
The nature of the relationships in both summaries is different: one is professional/collaborative, and the other is familial. \\ 
No\\
\\

$\blacktriangleright$\textbf{ Trained LLMs with \reward Output:} \\
Relation\_Summarization\_1:  \colorbox{nblue!35}{Sarah Kunstler is Emily's.}\\
Relation\_Summarization\_2:  He is Adam Lange's brother.\\
Understanding Process:  \\
In Relation\_Summarization\_1, the relation is familial, indicating that Sarah Kunstler and Emily are sisters.  \\
In Relation\_Summarization\_2, the relation is also \colorbox{nblue!35}{familial}, indicating that "he" is Adam Lange's \colorbox{nblue!35}{brother}.  \\
In both cases, the relation is the same: they are \colorbox{nblue!35}{siblings}.  \\
Yes\\

\bottomrule
\end{tabular}
\end{adjustbox}

\end{table*}

\begin{table*}[htbp]
\centering
\begin{adjustbox}{width=1.8\columnwidth,center}
\begin{tabular}{p{14cm} }
\hline

\rowcolor{Gainsboro} \multicolumn{1}{c}{\textit{Cases6 - Error Analysis from three Stages LLMs}} \\

$\blacktriangleright$\textbf{ Input:}\\
1.\textbf{China}'s National Development and Reform Commission (\textbf{NDRC}) on July 1 released a tape recording as evidence of the traders' mung bean price manipulation.\\
2. A US recession is likely to affect the industry, but strong growth in key Asian economies such as China and South Korea will drive regional travel demand, the \textbf{Pacific Asia Travel Association} (\textbf{PATA}) said.\\
\\

$\blacktriangleright$\textbf{ Gold Relation Labels and Ground Truth:}\\
1. org:member\_of \\
2. no\_relation \\
3. No \\
\\

$\blacktriangleright$\textbf{ Vanilla LLMs Output:} \\
Relation\_Summarization\_1:  \\
\colorbox{nyellow!90}{NDRC is an abbreviation for China's National Development and Reform Commission.}\\
Relation\_Summarization\_2:  PATA is an abbreviation for the Pacific Asia Travel Association.\\
The relations in both summaries involve an abbreviation: NDRC is an abbreviation for China's National Development and Reform Commission, and PATA is an abbreviation for the Pacific Asia Travel Association.\\  
Both summaries describe the relationship as one of \colorbox{nyellow!90}{abbreviation}. \\ 
Yes.\\
\\

$\blacktriangleright$\textbf{ Trained LLMs with Acc. Output:} \\
Relation\_Summarization\_1:  \\
\colorbox{nyellow!90}{NDRC is an abbreviation for China's National Development and Reform Commission.}

Relation\_Summarization\_2:  
PATA is an abbreviation for the Pacific Asia Travel Association.

Understanding Process:  
- In Relation\_Summarization\_1, the relation is that NDRC is an abbreviation for a specific organization within China.  
- In Relation\_Summarization\_2, the relation is that PATA is an abbreviation for a specific organization.  
- Both summaries describe an \colorbox{nyellow!90}{abbreviation} relationship between an \colorbox{nyellow!90}{acronym} and its full organizational name.  

Conclusion:  
Yes.\\
\\

$\blacktriangleright$\textbf{ Trained LLMs with \reward Output:} \\
Relation\_Summarization\_1:  \colorbox{nblue!35}{NDRC is China's National Development and Reform Commission.}\\
Relation\_Summarization\_2:  Pacific Asia Travel Association is abbreviated as PATA.\\

Understanding Process:  
In Relation\_Summarization\_1, the relation is organizational, indicating that NDRC is an \colorbox{nblue!35}{organization within} China.  
In Relation\_Summarization\_2, the relation is linguistic, indicating that PATA is an abbreviation for Pacific Asia Travel Association.  
The nature of these relations is different: one is about \colorbox{nblue!35}{organizational identity within} a country, and the other is about nomenclature.  

No\\

\bottomrule
\end{tabular}
\end{adjustbox}

\end{table*}

\clearpage
\newpage
\subsection{Comparison of Extracted Dictionaries by Open-source and Closed-source LLMs}\label{appendix:keywords_comparison}

We provide the complete phrase lists for all relation labels,
extracted separately by Qwen2.5-14B-Instruct and GPT-4o.
For each relation, we report the phrases identified by both models along with
representative explanation cases, enabling a detailed comparison of phrase
consistency across different LLMs.

\begin{table}[h]
\centering
\small
\begin{tabular}{p{\dimexpr0.4\linewidth-2\tabcolsep\relax} p{\dimexpr0.6\linewidth-2\tabcolsep\relax}}
\toprule
\textbf{Qwen2.5-14B} & \textbf{GPT4o} \\
\midrule
\rowcolor{Gainsboro}
\textit{/\allowbreak location/\allowbreak country/\allowbreak capital} &
\textit{/\allowbreak location/\allowbreak country/\allowbreak capital} \\
\midrule
\rowcolor{nyellow}
``city'' &
``city'' \\
\rowcolor{nyellow}
``capital'' &
``capital'' \\
\rowcolor{nyellow}
``location'' &
``location'' \\
\rowcolor{nyellow}
``government'' &
``government'' \\

``center'' & \\

``country'' &
\\

& ``city'' \\

\bottomrule
\end{tabular}
\vspace{5pt}
\caption{The phrases for the label
\colorbox{Gainsboro}{\textit{/location/country/capital}} extracted separately by Qwen2.5-14B and GPT-4o.
The \colorbox{nyellow}{yellow} indicates the same phrases generated by both
open-source and closed-source models.}
\end{table}

\begin{table}[h]
\centering
\small
\begin{tabular}{p{\dimexpr0.4\linewidth-2\tabcolsep\relax} p{\dimexpr0.6\linewidth-2\tabcolsep\relax}}
\toprule
\textbf{Qwen2.5-14B} & \textbf{GPT4o} \\
\midrule
\rowcolor{Gainsboro}
\textit{/\allowbreak location/\allowbreak location/\allowbreak contains} &
\textit{/\allowbreak location/\allowbreak location/\allowbreak contains} \\
\midrule
\rowcolor{nyellow}
``city'' &
``city'' \\
\rowcolor{nyellow}
``location'' &
``location'' \\
\rowcolor{nyellow}
``province'' &
``provincial'' \\
\rowcolor{nyellow}
``state'' &
``capital'' \\
\rowcolor{nyellow}
``located'' &
``located'' \\

``part'' &
 \\
 & ``contains'' \\

\bottomrule
\end{tabular}
\vspace{5pt}
\caption{The phrases for the label\colorbox{Gainsboro}{
\textit{/\allowbreak location/\allowbreak location/\allowbreak contains}} extracted separately by Qwen2.5-14B and GPT-4o.
The \colorbox{nyellow}{yellow} indicates the same phrases generated by both
open-source and closed-source models.}
\end{table}

\begin{table}[h]
\centering
\small
\begin{tabular}{p{\dimexpr0.4\linewidth-2\tabcolsep\relax} p{\dimexpr0.6\linewidth-2\tabcolsep\relax}}
\toprule
\textbf{Qwen2.5-14B} & \textbf{GPT4o} \\
\midrule
\rowcolor{Gainsboro}
\textit{/\allowbreak location/\allowbreak us\_\allowbreak state/\allowbreak capital} &
\textit{/\allowbreak location/\allowbreak us\_\allowbreak state/\allowbreak capital} \\
\midrule
\rowcolor{nyellow}
``city'' &
``city'' \\

\rowcolor{nyellow}
``located'' &
``located'' \\
\rowcolor{nyellow}
``capital'' &
``capital'' \\

``state'' & \\
& ``part''  \\

\bottomrule
\end{tabular}
\vspace{5pt}
\caption{The phrases for the label
\colorbox{Gainsboro}{\textit{/location/us\_state/capital}} extracted separately by Qwen2.5-14B and GPT-4o.
The \colorbox{nyellow}{yellow} indicates the same phrases generated by both
open-source and closed-source models.}
\end{table}

\begin{table}[h]
\centering
\small
\begin{tabular}{p{\dimexpr0.4\linewidth-2\tabcolsep\relax} p{\dimexpr0.6\linewidth-2\tabcolsep\relax}}
\toprule
\textbf{Qwen2.5-14B} & \textbf{GPT4o} \\
\midrule
\rowcolor{Gainsboro}
\textit{/\allowbreak business/\allowbreak company/\allowbreak place\_\allowbreak founded} &
\textit{/\allowbreak business/\allowbreak company/\allowbreak place\_\allowbreak founded} \\
\midrule
\rowcolor{nyellow}
``based'' &
``based'' \\
\rowcolor{nyellow}
``located'' &
``located'' \\
\rowcolor{nyellow}
``headquarters'' &
``headquarters'' \\
\rowcolor{nyellow}
``offices'' &
``offices'' \\
\rowcolor{nyellow}
``building'' &
``building'' \\
\rowcolor{nyellow}
``logo'' &
``logo'' \\

\bottomrule
\end{tabular}
\vspace{5pt}
\caption{The phrases for the label\colorbox{Gainsboro}{
\textit{/\allowbreak business/\allowbreak company/\allowbreak place\_\allowbreak founded}} extracted separately by Qwen2.5-14B and GPT-4o.
The \colorbox{nyellow}{yellow} indicates the same phrases generated by both
open-source and closed-source models.}
\end{table}

\begin{table}[h]
\centering
\small
\begin{tabular}{p{\dimexpr0.4\linewidth-2\tabcolsep\relax} p{\dimexpr0.6\linewidth-2\tabcolsep\relax}}
\toprule
\textbf{Qwen2.5-14B} & \textbf{GPT4o} \\
\midrule
\rowcolor{Gainsboro}
\textit{/\allowbreak film/\allowbreak film\_\allowbreak location/\allowbreak featured\_\allowbreak in\_\allowbreak films} &
\textit{/\allowbreak film/\allowbreak film\_\allowbreak location/\allowbreak featured\_\allowbreak in\_\allowbreak films} \\
\midrule
\rowcolor{nyellow}
 ``setting'' &
``setting''\\
\rowcolor{nyellow}
 ``set'' &
``set''\\
\rowcolor{nyellow}
 ``place'' &
``place''\\
\rowcolor{nyellow}
 ``associated'' &
``associated''\\

\bottomrule
\end{tabular}
\vspace{5pt}
\caption{The phrases for the label\colorbox{Gainsboro}{
\textit{/\allowbreak film/\allowbreak film\_\allowbreak location/\allowbreak featured\_\allowbreak in\_\allowbreak films}} extracted separately by Qwen2.5-14B
and GPT-4o. The \colorbox{nyellow}{yellow} indicates the same phrases generated by
both open-source and closed-source models.}
\end{table}


\begin{table}[h]
\centering
\small
\begin{tabular}{p{\dimexpr0.4\linewidth-2\tabcolsep\relax} p{\dimexpr0.6\linewidth-2\tabcolsep\relax}}
\toprule
\textbf{Qwen2.5-14B} & \textbf{GPT4o} \\
\midrule
\rowcolor{Gainsboro}
\textit{/\allowbreak business/\allowbreak company/\allowbreak founders} &
\textit{/\allowbreak business/\allowbreak company/\allowbreak founders} \\
\midrule
\rowcolor{nyellow}
 ``chairman'' &
``chairman''\\
\rowcolor{nyellow}
 ``founder'' &
``founder''\\
\rowcolor{nyellow}
 ``associated'' &
``associated''\\
\rowcolor{nyellow}
 ``chief'' &
``chief''\\
\rowcolor{nyellow}
 ``officer'' &
``officer''\\
\rowcolor{nyellow}
 ``executive'' &
``executive''\\
\rowcolor{nyellow}
 ``co-founder'' &
``co-founder''\\
\rowcolor{nyellow}
 ``president'' &
``president''\\

 ``member'' &
\\

\bottomrule
\end{tabular}
\vspace{5pt}
\caption{The phrases for the label\colorbox{Gainsboro}{
\textit{/\allowbreak business/\allowbreak company/\allowbreak founders}} extracted separately by Qwen2.5-14B and GPT-4o.
The \colorbox{nyellow}{yellow} indicates the same phrases generated by
both open-source and closed-source models.}
\end{table}

\begin{table}[h]
\centering
\small
\begin{tabular}{p{\dimexpr0.4\linewidth-2\tabcolsep\relax} p{\dimexpr0.6\linewidth-2\tabcolsep\relax}}
\toprule
\textbf{Qwen2.5-14B} & \textbf{GPT4o} \\
\midrule
\rowcolor{Gainsboro}
\textit{/\allowbreak people/\allowbreak person/\allowbreak nationality} &
\textit{/\allowbreak people/\allowbreak person/\allowbreak nationality} \\
\midrule
\rowcolor{nyellow}
 ``President'' &
``president''\\
\rowcolor{nyellow}
 ``Minister'' &
``minister''\\
\rowcolor{nyellow}
 ``Prime'' &
``prime''\\
\rowcolor{nyellow}
 ``chancellor'' &
``chancellor''\\

\bottomrule
\end{tabular}
\vspace{5pt}
\caption{The phrases for the label\colorbox{Gainsboro}{
\textit{/\allowbreak people/\allowbreak person/\allowbreak nationality}} extracted separately by Qwen2.5-14B and GPT-4o.
The \colorbox{nyellow}{yellow} indicates the same phrases generated by
both open-source and closed-source models.}
\end{table}

\begin{table}[h]
\centering
\small
\begin{tabular}{p{\dimexpr0.4\linewidth-2\tabcolsep\relax} p{\dimexpr0.6\linewidth-2\tabcolsep\relax}}
\toprule
\textbf{Qwen2.5-14B} & \textbf{GPT4o} \\
\midrule
\rowcolor{Gainsboro}
\textit{/\allowbreak business/\allowbreak person/\allowbreak company} &
\textit{/\allowbreak business/\allowbreak person/\allowbreak company} \\
\midrule
\rowcolor{nyellow}
 ``Secretary'' &
``secretary'' \\
\rowcolor{nyellow}
 ``President'' &
``president''\\
\rowcolor{nyellow}
 ``program'' &
``program''\\
\rowcolor{nyellow}
 ``executive'' &
``executive''\\
\rowcolor{nyellow}
 ``chief'' &
``chief''\\
\rowcolor{nyellow}
 ``leader'' &
``leader''\\
\rowcolor{nyellow}
 ``chairman'' &
``chairman''\\
\rowcolor{nyellow}
 ``officer'' &
``officer''\\
\rowcolor{nyellow}
 ``correspondent'' &
``correspondent''\\

\bottomrule
\end{tabular}
\vspace{5pt}
\caption{The phrases for the label
\colorbox{Gainsboro}{\textit{/business/person/company}} extracted separately by Qwen2.5-14B and GPT-4o.
The \colorbox{nyellow}{yellow} indicates the same phrases generated by both
open-source and closed-source models.}
\end{table}

\begin{table}[h]
\centering
\small
\begin{tabular}{p{\dimexpr0.4\linewidth-2\tabcolsep\relax} p{\dimexpr0.6\linewidth-2\tabcolsep\relax}}
\toprule
\textbf{Qwen2.5-14B} & \textbf{GPT4o} \\
\midrule
\rowcolor{Gainsboro}
\textit{/\allowbreak location/\allowbreak us\_\allowbreak county/\allowbreak county\_\allowbreak seat} &
\textit{/\allowbreak location/\allowbreak us\_\allowbreak county/\allowbreak county\_\allowbreak seat} \\
\midrule
\rowcolor{nyellow}
 ``municipality'' &
``municipality''\\
\rowcolor{nyellow}
 ``included'' &
``included''\\
\rowcolor{nyellow}
 ``area'' &
``area''\\
\rowcolor{nyellow}
 ``around'' &
``around''\\
\rowcolor{nyellow}
 ``in'' &
``in''\\
\rowcolor{nyellow}
 ``part'' &
``part''\\

\bottomrule
\end{tabular}
\vspace{5pt}
\caption{The phrases for the label
\colorbox{Gainsboro}{\textit{/location/us\_county/county\_seat}} extracted separately by Qwen2.5-14B and GPT-4o.
The \colorbox{nyellow}{yellow} indicates the same phrases generated by both
open-source and closed-source models.}
\end{table}

\begin{table}[h]
\centering
\small
\begin{tabular}{p{\dimexpr0.4\linewidth-2\tabcolsep\relax} p{\dimexpr0.6\linewidth-2\tabcolsep\relax}}
\toprule
\textbf{Qwen2.5-14B} & \textbf{GPT4o} \\
\midrule
\rowcolor{Gainsboro}
\textit{/\allowbreak people/\allowbreak person/\allowbreak place\_\allowbreak lived} &
\textit{/\allowbreak people/\allowbreak person/\allowbreak place\_\allowbreak lived} \\
\midrule
\rowcolor{nyellow}
 ``lives'' &
``lives''\\
\rowcolor{nyellow}
 ``hometown'' &
``hometown''\\
\rowcolor{nyellow}
 ``from'' &
``from''\\
\rowcolor{nyellow}
 ``resides'' &
``resides''\\
\rowcolor{nyellow}
 ``governor'' &
``governor''\\

 ``of'' & \\
 & ``at''\\
\bottomrule
\end{tabular}
\vspace{5pt}
\caption{The phrases for the label
\colorbox{Gainsboro}{\textit{/people/person/place\_lived}} extracted separately by Qwen2.5-14B and GPT-4o.
The \colorbox{nyellow}{yellow} indicates the same phrases generated by both
open-source and closed-source models.}
\end{table}

\begin{table}[h]
\centering
\small
\begin{tabular}{p{\dimexpr0.4\linewidth-2\tabcolsep\relax} p{\dimexpr0.6\linewidth-2\tabcolsep\relax}}
\toprule
\textbf{Qwen2.5-14B} & \textbf{GPT4o} \\
\midrule
\rowcolor{Gainsboro}
\textit{/\allowbreak people/\allowbreak person/\allowbreak children} &
\textit{/\allowbreak people/\allowbreak person/\allowbreak children} \\
\midrule
\rowcolor{nyellow}
 ``mother'' &
``mother''\\
\rowcolor{nyellow}
 ``son'' &
``son''\\
\rowcolor{nyellow}
 ``father'' &
``father'' \\
\rowcolor{nyellow}
 ``daughter'' &
``daughter''\\
\rowcolor{nyellow}
 ``children'' &
``child''\\

 & ``niece'' \\

 & ``relative''\\

\bottomrule
\end{tabular}
\vspace{5pt}
\caption{The phrases for the label
\colorbox{Gainsboro}{\textit{/people/person/children}} extracted separately by Qwen2.5-14B and GPT-4o.
The \colorbox{nyellow}{yellow} indicates the same phrases generated by both
open-source and closed-source models.}
\end{table}

\begin{table}[h]
\centering
\small
\begin{tabular}{p{\dimexpr0.4\linewidth-2\tabcolsep\relax} p{\dimexpr0.6\linewidth-2\tabcolsep\relax}}
\toprule
\textbf{Qwen2.5-14B} & \textbf{GPT4o} \\
\midrule
\rowcolor{Gainsboro}
\textit{/\allowbreak people/\allowbreak ethnicity/\allowbreak geographic\_\allowbreak distribution} &
\textit{/\allowbreak people/\allowbreak ethnicity/\allowbreak geographic\_\allowbreak distribution} \\
\midrule
\rowcolor{nyellow}
 ``energy'' &
``energy''\\
\rowcolor{nyellow}
 ``gas'' &
``gas''\\
\rowcolor{nyellow}
 ``loan'' &
``loan''\\
\rowcolor{nyellow}
 ``reporter'' &
``reporter'' \\
\rowcolor{nyellow}
 ``national'' &
``national''\\
\rowcolor{nyellow}
 ``associated'' &
``associated''\\
\rowcolor{nyellow}
 ``pipelines'' &
``pipelines''\\

 ``consortium'' &
\\

 ``identity'' &
\\

 & ``foreign''\\

\bottomrule
\end{tabular}
\vspace{5pt}
\caption{The phrases for the label\colorbox{Gainsboro}{
\textit{/\allowbreak people/\allowbreak ethnicity/\allowbreak geographic\_\allowbreak distribution}} extracted separately by
Qwen2.5-14B and GPT-4o. The \colorbox{nyellow}{yellow} indicates the same phrases
generated by both open-source and closed-source models.}
\end{table}

\begin{table}[h]
\centering
\small
\begin{tabular}{p{\dimexpr0.4\linewidth-2\tabcolsep\relax} p{\dimexpr0.6\linewidth-2\tabcolsep\relax}}
\toprule
\textbf{Qwen2.5-14B} & \textbf{GPT4o} \\
\midrule
\rowcolor{Gainsboro}
\textit{/\allowbreak people/\allowbreak person/\allowbreak ethnicity} &
\textit{/\allowbreak people/\allowbreak person/\allowbreak ethnicity} \\
\midrule
\rowcolor{nyellow}
 ``American'' &
``american''\\
\rowcolor{nyellow}
 ``Italian'' &
``italian''\\
\rowcolor{nyellow}
 ``Manchu'' &
``manchu''\\
\rowcolor{nyellow}
 ``Russian'' &
``russian''\\
\rowcolor{nyellow}
 ``Tejano'' &
``tejano''\\

\bottomrule
\end{tabular}
\vspace{5pt}
\caption{The phrases for the label\colorbox{Gainsboro}{
\textit{/\allowbreak people/\allowbreak person/\allowbreak ethnicity}} extracted separately by Qwen2.5-14B and GPT-4o.
The \colorbox{nyellow}{yellow} indicates the same phrases generated by both
open-source and closed-source models.}
\end{table}

\begin{table}[h]
\centering
\small
\begin{tabular}{p{\dimexpr0.4\linewidth-2\tabcolsep\relax} p{\dimexpr0.6\linewidth-2\tabcolsep\relax}}
\toprule
\textbf{Qwen2.5-14B} & \textbf{GPT4o} \\
\midrule
\rowcolor{Gainsboro}
\textit{/\allowbreak people/\allowbreak deceased\_\allowbreak person/\allowbreak place\_\allowbreak of\_\allowbreak burial} &
\textit{/\allowbreak people/\allowbreak deceased\_\allowbreak person/\allowbreak place\_\allowbreak of\_\allowbreak burial} \\
\midrule
\rowcolor{nyellow}
 ``home'' &
``home''\\

 & ``sites''\\
\rowcolor{nyellow}
 ``buried'' &
``buried''\\

 & ``grave'',\\
\rowcolor{nyellow}
 ``interred'' &
``interred'' \\
\rowcolor{nyellow}
 ``cemetery'' &
``cemetery'' \\

\bottomrule
\end{tabular}
\vspace{5pt}
\caption{The phrases for the label\colorbox{Gainsboro}{
\textit{/\allowbreak people/\allowbreak deceased\_\allowbreak person/\allowbreak place\_\allowbreak of\_\allowbreak burial}} extracted separately by
Qwen2.5-14B and GPT-4o. The \colorbox{nyellow}{yellow} indicates the same phrases
generated by both open-source and closed-source models.}
\end{table}

\begin{table}[h]
\centering
\small
\begin{tabular}{p{\dimexpr0.4\linewidth-2\tabcolsep\relax} p{\dimexpr0.6\linewidth-2\tabcolsep\relax}}
\toprule
\textbf{Qwen2.5-14B} & \textbf{GPT4o} \\
\midrule
\rowcolor{Gainsboro}
\textit{/\allowbreak people/\allowbreak place\_\allowbreak of\_\allowbreak interment/\allowbreak interred\_\allowbreak here} &
\textit{/\allowbreak people/\allowbreak place\_\allowbreak of\_\allowbreak interment/\allowbreak interred\_\allowbreak here} \\
\midrule
\rowcolor{nyellow}
 ``home'' &
``home''\\
\rowcolor{nyellow}
 ``tomb'' &
``tomb''\\
\rowcolor{nyellow}
 ``buried'' &
``buried''\\
\rowcolor{nyellow}
 ``site'' &
``sites''\\
\rowcolor{nyellow}
 ``staging'' &
``staging''\\
\rowcolor{nyellow}
 ``rites'' &
``rites''\\

 & ``birthplace''\\
\rowcolor{nyellow}
 ``residence'' &
``resided''\\

\bottomrule
\end{tabular}
\vspace{5pt}
\caption{The phrases for the label
\colorbox{Gainsboro}{\textit{/people/place\_of\_interment/interred\_here}} extracted separately by
Qwen2.5-14B and GPT-4o. The \colorbox{nyellow}{yellow} indicates the same phrases
generated by both open-source and closed-source models.}
\end{table}

\end{document}